\documentclass[mnsc,sglanonrev]{informs4-nored}
\usepackage{eqndefns-left} % For checking the display equation width and equation environment definitions %
\RequirePackage{tgtermes}
\RequirePackage{newtxtext}
\RequirePackage{newtxmath}
\RequirePackage{bm}
\RequirePackage{endnotes}

\OneAndAHalfSpacedXII % Current default line spacing
\usepackage{algorithm}
\usepackage{algpseudocode}
\usepackage{tikz}
\usepackage{booktabs}  % 用于三线表的宏包
\usepackage{xcolor}      % 用于背景色
\usepackage{enumitem}

\usepackage{natbib}
 \bibpunct[, ]{(}{)}{,}{a}{}{,}%
 \def\bibfont{\small}%

\usepackage[hidelinks]{hyperref}
\hypersetup{
    colorlinks,
    breaklinks,
    linkcolor=blue,
    urlcolor=black,
    anchorcolor=black,
    citecolor=blue
}

\EquationsNumberedThrough    % Default: (1), (2), ...
\TheoremsNumberedThrough     % Preferred (Theorem 1, Lemma 1, Theorem 2)
\ECRepeatTheorems  %  

\MANUSCRIPTNO{MNSC-0001-2024.00}

\begin{document}
%%%%%%%%%%%%%%%%

%\setcounter{page}{1}
%\DOI{10.1287/xxxx.0000.0000}%

\RUNAUTHOR{Zhao et al.}

% \RUNTITLE{Predictive Maintenance in Manufacturing}
\RUNTITLE{OR-Guided Pretrain-then-Reinforce Framework for Inventory Management}

% \TITLE{Optimal Resource Allocation in Humanitarian Logistics: A Stochastic Programming Approach}
\TITLE{ORPR: An OR-Guided Pretrain-then-Reinforce Learning Model for Inventory Management}

\ARTICLEAUTHORS{%
% 作者名合并为一行，单位用字母上标标注
\AUTHOR{Lingjie Zhao$^{\text{a},\dagger}$, Xue Yu$^{\text{b},\dagger}$, Yongzhi Qi$^\text{b,*}$, Hao Hu$^\text{b}$, Jianshen Zhang$^\text{b}$, Yingzheng Ma$^\text{b}$, Shuyu Han$^\text{b}$, Wei Qi$^\text{a,*}$, Zuo-Jun Max Shen$^\text{c}$}

% 单位合并为一行，对应字母标注
\AFF{$^\text{a}$Department of Industrial Engineering, Tsinghua University, Beijing 100084, China; $^\text{b}$Supply Chain Tech Team Y, JD.com, Beijing 101111, China; $^\text{c}$Faculty of Engineering and Faculty of Business and Economics, The University of Hong Kong, Hong Kong 999077, China}

\AFF{$^{*}$Corresponding author; $^\dagger$Equal Contribution.}
% 联系方式（合并为一行，包含邮箱+ORCID）
\AFF{\textbf{Contact:} zhaolj25@mails.tsinghua.edu.cn (LZ); yuxue31@jd.com (XY); qiyongzhi1@jd.com (YQ); huhao@jd.com (HH); zhangjianshen@jd.com (JZ); mayingzheng.1@jd.com (YM); hanshuyu@jd.com (SH); qiw@tsinghua.edu.cn (WQ); maxshen@hku.hk (Z-JMS)}
} % end of the block

\ABSTRACT{%
As the pursuit of synergy between Artificial Intelligence (AI) and Operations Research (OR) gains momentum in handling complex inventory systems, a critical challenge persists: how to effectively reconcile AI's adaptive perception with OR's structural rigor. To bridge this gap, we propose a novel OR-Guided ``Pretrain-then-Reinforce" framework. To provide structured guidance, we propose a simulation-augmented OR model that generates high-quality reference decisions, implicitly capturing complex business constraints and managerial preferences. Leveraging these OR-derived decisions as foundational training labels, we design a domain-informed deep learning foundation model to establish foundational decision-making capabilities, followed by a reinforcement learning (RL) fine-tuning stage. Uniquely, we position RL as a deep alignment mechanism that enables the AI agent to internalize the optimality principles of OR, while simultaneously leveraging exploration for general policy refinement and allowing expert guidance for scenario-specific adaptation (e.g., promotional events). Validated through extensive numerical experiments and a field deployment at JD.com augmented by a Difference-in-Differences (DiD) analysis, our model significantly outperforms incumbent industrial practices, delivering real-world gains of a 5.27-day reduction in turnover and a 2.29\% increase in in-stock rates, alongside a 29.95\% decrease in holding costs. Contrary to the prevailing trend of brute-force model scaling, our study demonstrates that a lightweight, domain-informed model can deliver state-of-the-art performance and robust transferability when guided by structured OR logic. This approach offers a scalable and cost-effective paradigm for intelligent supply chain management, highlighting the value of deeply aligning AI with OR.

}%

% \FUNDING{This research was supported by [grant number, funding agency].}

%Supplemental Material:
%Data Ethics & Reproducibility Note:

%\KEYWORDS{Stochastic programming, Decision support,Uncertainty, Disaster response, Optimization}

%\HISTORY{Received: Month DD, YYYY; Accepted: Month DD, YYYY; Published Online: Month DD, YYYY}

\maketitle
%%%%%%%%%%%%%%%%%%%%%%%%%%%%%%%%%%%%%%%%%%%%%%%%%%%%%%%%%%%%%%%%%%%%%%

% Text of your paper here
\section{Introduction}
Recent years have witnessed the confluence of two transformative trends:  the rapid advancement of artificial intelligence (AI) and the escalating complexity of real-world operational systems. Advanced AI systems, powered by scaling laws and ever-larger models, are profoundly expanding the frontier of data-driven decision-making.  Exhibiting impressive generalization capabilities across diverse tasks and domains, including urban management, smart manufacturing, drug discovery and operational data analytics (\citealp{li2025ed2mol,ren2025indust,yuan2024unist}), AI systems offer an unprecedented capacity to process high-dimensional data, capture nonlinear dependencies, and generate adaptive strategies. Nevertheless, their lack of transparency and domain-specific guarantees raises growing concerns about reliability, stability, and interpretability, posing significant challenges for their deployment in high-stakes practical applications.
Concurrently, modern operational environments are evolving into systems of unprecedented scale and intricacy. This trend places increasing strain on operations research (OR) methods, which have historically served as the analytical foundation for such systems. While OR provides clarity, interpretability, and optimality under well-defined assumptions, the very assumptions that grant OR its analytical elegance, such as stationarity and convexity, render its models increasingly ill-equipped to handle the complexity of modern systems.
Large-scale inventory management, which serves as the quintessential embodiment of this complexity, is characterized by competing multiple objectives, a wide portfolio of interdependent SKUs, and fine-grained business constraints interwoven with implicit managerial preferences. This reality creates a widening gap between theoretical optimality and practical relevance.

This confluence of trends brings our community to a critical juncture.
As supply chain management scholars and practitioners, we ponder what role OR methods can assume in this ongoing movement toward intelligent, data-driven inventory management. The ideal, we argue, is to forge a synergy that combines the adaptive generalization of AI with the analytical, structural rigor of OR. This philosophy of synergistic integration was presciently articulated by \cite{simon1987two}:

\begin{quote}
    we can apply an amalgamation of AI and OR analytic and problem-solving methods. Optimization methods can be applied to subproblems within larger environments that are too complex for them. Exact theoretical analyses of simplified problems can suggest procedures ... when applied to complex real world situations.
\end{quote}

Can the marriage of AI and OR truly realize this synergistic potential for inventory management? Until recently, the integration of OR and AI for inventory management remained in its infancy. To leverage the increasing availability of observable features and data, early efforts primarily followed the ``predict-then-optimization" paradigm. While effective in structured settings, such separation between prediction and decision often leads to a mismatch between the training objective and the actual decision performance. Moreover, relying on point estimates inevitably leads to information loss, while distributional forecasting remains highly challenging in practice. The heterogeneity, interdependence, and complex dynamics of demand time-series make such estimation at arbitrary horizons both inefficient and unreliable for our large-scale setting.

In response, recent research increasingly adopts end-to-end approaches that leverage deep learning (DL) models to directly map input features to replenishment decisions (\citealp{qi2023practical,liu2023ai}). However, these models often overfit historical patterns and short-term noise, resulting in poor generalization to out-of-distribution or long-tail demand scenarios. Another major line of work employs reinforcement learning (RL) to directly learn replenishment policies (\citealp{gijsbrechts2022can,harsha2025deep,mao2025model,xie2025DeepStock}). Nevertheless, the application of RL to inventory management is fraught with practical challenges, including high training costs, dependence on accurate simulators, and instability in large-scale state–action spaces \citep{levine2020offline,gijsbrechts2025AI}. Crucially, the vast majority of these studies remain confined to single-SKU settings, with limited exploration of multi-category joint replenishment, leaving a persistent gap between academic models and industrial-scale applications.

From a broader perspective, a growing body of research has provided valuable insights into integrating advanced AI with operations research (OR). \cite{elmachtoub2022smart} introduces the Smart Predict-then-Optimize (SPO) framework, which bridges the traditional divide between prediction and optimization by embedding the downstream decision-loss directly into the training loss. This approach demonstrates the added value of leveraging OR domain knowledge to guide predictive learning for better decision outcomes. Recent work also explores fine-tuning and calibration of large foundation models for solving optimization problems, supported by effective pipelines for generating (problem–solution) pairs, highlighting the growing potential of AI and large models in advancing decision-making and optimization (\citealp{huang2025orlm,zhou2025dplm}). However, the application of such approaches to inventory management remains largely underexplored.

Having observed the pursuits of integrating OR and machine learning, we feel that it is timely and relevant to develop a new framework specifically architected for inventory management, which moves beyond simply applying existing AI architectures. Informed by our domain expertise and practical experience, we express our initial thoughts in this paper by addressing two foundational questions that guide our methodology.
\begin{itemize}
    \item \textit{What are the distinct and complementary roles of AI and OR in a unified decision-making system?} We contend that AI should serve as the engine for perception and adaptation, responsible for interpreting complex signals from a high-dimensional environment. Conversely, OR's role is to provide the structural logic, encoding core economic principles and operational constraints to define a structured and feasible decision space.
    \item \textit{What is the most effective architecture for synergy?} We posit that the most effective integration transcends simple sequential pipelines (like PTO) or the mere fitting of OR labels via supervised learning. Instead, we advocate for a deep alignment architecture facilitated by reinforcement learning. In this paradigm, RL serves as a mechanism to establish a profound connection between AI and OR, enabling the AI to internalize the decision patterns and optimality principles of OR, while leveraging exploration to drive continuous decision improvement.
\end{itemize}

% (tentative is ok? seems a little weak)
Building on these tentative thoughts, we develop an OR-guided ``pre-train + RL fine-tune" framework for large-scale inventory management in collaboration with JD.com, China's largest e-commerce platform by revenue. The framework operates in two stages: first, a deep learning model is pre-trained to generate strong feature representations and a high-quality baseline policy. Second, reinforcement learning fine-tunes this policy, aligning its decisions with OR principles and domain knowledge while exploring the policy space to discover superior strategies. For guiding the training phase, we propose a simulation-augmented OR model to generate reference decisions. This simulation process unfolds over a given time horizon to evaluate replenishment decisions within a realistic operational setting, taking as input historical demand realizations and empirical parameters including stockout penalties, holding costs, profit margins and vendor lead times. For any given replenishment policy, the simulation outputs key performance metrics, including realized stockout costs and end-of-period inventory levels, which serve as crucial parameters for our OR model. This simulation-augmented approach transcends the limitations of explicit analytical modeling, which can be intractable and prone to misspecification. This allows us to generate reference decisions that more closely approximate the true operational optimum. The resulting OR-guided Pretrain-then-Reinforce model achieves an efficient, scalable, and customizable solution for large-scale, heterogeneous inventory decision management problems, achieving state-of-the-art performance without requiring a vast model scale.

% %(is the first sentance too informal?) 
This proposed model turns out to be a success. Since September 2025, our model has been deployed in JD.com’s live retail environment, managing inventory for a subset of SKUs across three categories, including ``meat snacks," ``cakes \& pastries" and ``egg snacks." We conducted a field experiment from September 1, 2025 to October 1, 2025. The experiment involved 3,899 distinct (SKU-DC) pairs for 331 SKUs. The experiment compares the performance of our OR-guided Pretrain-then-Reinforce model against JD.com's incumbent replenishment algorithm. The results show that the OR-guided Pretrain-then-Reinforce model dominates the current algorithm across all performance measures. More specifically, the holding cost for the treatment group is reduced by about 30\% compared with the control group, whereas the inventory turnover decreases by 5.27 days and the  in-stock rate increases by 2.29\%. We also adopt a difference-in-difference (DiD) approach to isolate the time effects. The field experiment results demonstrate the immediate applicability and advantages of our proposed OR-guided Pretrain-then-Reinforce approach. 
% (最后这句是否call back一下，例如These field results provide strong empirical evidence for the superiority of our framework, demonstrating that a structured integration of OR and AI can successfully bridge the gap between theoretical models and the complex demands of large-scale industrial practice.)

Our contributions can be summarized in three main aspects:
\begin{enumerate}
    \item \textit{A New Framework for AI+OR in Inventory Management.} We propose a novel OR-Guided ``Pretrain-then-Reinforce" framework that integrates a domain-informed deep neural network pretraining with Reinforcement Learning fine-tuning. We also formulate a simulation-augmented OR model to generate reference decisions for guiding the training phases. 
    To the best of our knowledge, this represents the first practical application of a pre-training + RL fine-tuning paradigm for inventory management, and we believe it offers a general pathway to align the learning and generalization capacity of AI with the optimality principles and domain understanding of OR. 
    \item \textit{The Critical Role of OR in AI-driven Decisions.}  We show that the role of OR extends beyond solving subproblems to actively guiding the entire AI learning process. By using a structured OR model to guide the learning process of the AI agent, AI systems can achieve superior performance and alignment with business objectives and optimal pattern. Our work highlights the efficiency of this domain-aware design, which achieves these results with a model of only a few million parameters, providing a nimble alternative to resource-intensive, billion-parameter foundation models.
    \item \textit{Real-world efficacy.} In the numerical experiments, our approach outperforms multiple baseline models, including JD.com’s current operational practice, across both offline simulations based on real-world data and a field experiment. A case study using real data further demonstrates the value of RL in aligning the model outputs with dynamically evolving business requirements in critical scenarios such as major promotional events. The field experiment further substantiates the practical applicability of our model in real-world e-commerce inventory management.
\end{enumerate}

The remainder of this paper is organized as follows. Section~\ref{sec:literature} reviews the relevant literature. Section~\ref{sec:models} describes the components of our framework, including the multi-category joint replenishment decision OR model, the deep learning foundation model, and the reinforcement learning fine‑tuning algorithm. In Section~\ref{sec:num_exp}, we detail our numerical experiments and report the results. Section~\ref{sec:field_exp} presents findings from the field experiment conducted at JD.com. 
Finally, we conclude in Section~\ref{sec:conclusion} and discuss potential future research directions.
\section{Literature Review}
\label{sec:literature}
Inventory management has been a subject of extensive scholarly inquiry for decades. In his seminal work, \cite{scarf1960optimality} examine the optimal replenishment policy for a single supplier under infinite supply, constant lead time, and fixed ordering costs, establishing the classic $(s,S)$ structure. \cite{zipkin1984efficient} propose an efficient algorithm under standard assumptions for computing the optimal $(s,S)$ policy. \cite{ehrhardt1984s} extend this framework to multi-period inventory systems with stochastic lead times and proves optimality under certain conditions. In lost-sales settings, \cite{zipkin2008old} demonstrate that even the single-supplier problem admits no closed-form characterization of the optimal policy structure. Subsequent research further generalizes these models, developing a diverse array of parametric inventory policies tailored to various operational contexts. For a comprehensive review on this topic, we recommend textbooks such as \cite{zipkin2000foundations} and \cite{snyder2019fundamentals}.

However, these methods often fail to remain effective in large-scale, real-world applications characterized by factors such as nonstationary demand. Therefore, industry practitioners commonly employ a two-step predict-then-optimize (PTO) framework: First, forecast demand and other uncertainties, then embed those forecasts into an optimization model or decision rule, such as the classical $(s,S)$ policy. Forecasting outputs may take the form of point estimates or full predictive distributions, the latter of which has been extensively studied and deployed in practice (\citealp{bertsimas2020predictive,sheather1991reliable}). Although PTO's modularity and ease of implementation make it attractive for practitioners, recent studies have demonstrated that decoupling prediction and optimization will amplify even small forecasting errors during the optimization phase, leading to significant decision biases and out-of-sample performance degradation (\citealp{smith2006optimizer,ramamurthy2012inventory,liu2021time}).

To mitigate this issue, recent data-driven optimization research has pursued two main directions: integrating prediction and decision in a unified model (\citealp{elmachtoub2022smart,mandi2022decision}), and employing end-to-end learning (\citealp{qi2023practical,harsha2025deep}). Within the newsvendor context, a number of works have explored feature-based inventory policies. \cite{ban2019big} develop a unified optimization framework that directly leverages historical demand features to determine order quantities, while \cite{oroojlooyjadid2020applying} employ machine-learning ensembles to approximate the optimal order policy. However, these approaches typically target simplified settings and struggle to scale effectively to the high-dimensional, heterogeneous environments encountered in large-scale inventory management practice.

With the advent of deep learning, end-to-end approaches have emerged as a compelling alternative to mitigate the performance loss inherent in predict-then-optimize frameworks, and their application to inventory management has grown rapidly. \cite{qi2023practical} propose a deep neural network that parallelizes demand and lead-time-trigger (VLT) forecasting modules before concatenating them to directly predict replenishment actions; their approach delivers superior results in both numerical simulations and field trials at JD.com. Similarly, reinforcement learning methods, with their superior capabilities in reward learning and exploration, are increasingly being applied to tackle complex, high-dimensional dynamic problems. For instance, \cite{oroojlooyjadid2022deep} demonstrate the effectiveness of RL in the canonical Beer Game, where competing agents replenish the limited beer stock in a serial supply chain. In the context of dual-sourcing and lost-sales systems,  \cite{gijsbrechts2022can} and \cite{tang2024online} develop effecient learning-based policies to manage replenishment dynamics, showing performance advantage over dynamic-programming benchmarks. \cite{chen2022dynamic,chen2024optimal} extend the learning framework to the joint optimization of pricing and inventory control under incomplete information. \cite{gong2024bandits} model cycle-driven demand in zero-lead-time and positive-lead-time environments and show RL’s superiority on real sales data. Finally, \cite{boute2022deep} provide a comprehensive survey of RL applications in inventory management.

While these studies demonstrate the efficacy of learning-based approaches within their respective experimental domains, they often suffer from inherent flaws that can lead to significant losses in practical deployment. For example, many end-to-end deep learning (DL) methods still exhibit a fundamental mismatch between the training loss and the final decision objective, a disparity that has been shown in the literature to cause biased decision-making (\citealp{wilder2019melding,donti2017task}). Moreover, the policy outputs learned by the model can diverge from human preferences or the canonical optimal policy, and often lack interpretability. Concurrently, traditional enumeration-based reinforcement learning (RL) techniques are often computationally infeasible due to the large combinatorial action spaces with state-dependent constraints that are inherent to operations management (OM) problems like inventory management (\citealp{harsha2025deep}). 

Furthermore, the dependent, heterogeneous, and time-varying nature of demand poses a significant challenge to multi-category joint inventory decision-making. Most existing research focuses exclusively on single-SKU inventory decisions, failing to adequately account for the impact of complex substitution, competition, and other inter-product relationships on replenishment policy (\citealp{qi2023practical, harsha2025deep, gijsbrechts2022can}). Only a few studies address holistic inventory optimization across multiple product categories, and these efforts remain limited in scale (\citealp{meisheri_scalable_2022, liu2023deep}). Additionally, in the volatile and complex commercial environment, current models often lack the agility required for rapid iteration and adaptation to emergent demand patterns and evolving business objectives. These gaps collectively highlight the urgent need for highly adaptive, multi-category joint inventory optimization models.

Insights into solving these alignment and adaptation challenges can be drawn from a parallel domain: large language models (LLMs) (\citealp{huang2025orlm,zhou2025dplm}). The rapid development of LLMs has demonstrated the power of the ``pre-training + fine-tuning from feedback" paradigm. Concurrently, a growing body of recent work has begun to explore reinforcement learning (RL) not merely as a policy-learning algorithm, but as a post-training mechanism for output alignment and calibration (\citealp{ziegler2020finetuning, ji2025aialign}). One prominent example is Deepseek-R1, reported in \cite{guo2025deepseek} in \textit{Nature}, which employs RL to align the model's reasoning process with verifiable ground-truth outcomes, enabling the model to develop advanced behaviors like self-reflection.
% actually, RL in this paper seems to be a pure policy learning approach, while it uses no CoT data.
Unlike traditional RL methods that learn policies from scratch, RL-based fine-tuning in LLM post-training adjusts model weights according to preference feedback, enhancing response consistency and reasoning. Its key advantage lies in dynamically adapting to diverse output preferences rather than optimizing toward a single fixed target. Early RLHF  directly employed classical reinforcement learning algorithms such as Proximal Policy Optimization (PPO) (\citealp{ouyang2022training,bai2022constitutional}). Subsequent studies observed that a full RL architecture is often unnecessary for alignment tasks, as the reward signal is already implicit in the preference data. This realization gives rise to Direct Preference Optimization (DPO) and its improvements (\citealp{rafailov2023direct,meng2024simpo,ethayarajh2024kto}).
However, offline DPO-style methods sacrifice the exploration capability inherent in reinforcement learning, which has motivated subsequent studies to develop approaches such as RLOO and GRPO that remove the critic while preserving online exploration (\citealp{ahmadian2024back,shao2024deepseekmath}).
For a more detailed discussion of this literature, readers are directed to the comprehensive review by \cite{wang2024comprehensive}. By eliminating the need to learn an explicit reward model, these methods achieve higher sample efficiency, leading to reduced training costs and more robust fine-tuning.

Inspired by these advancements, our approach uniquely integrates a pre-trained deep learning (DL) model, reinforcement learning (RL) from feedback, and the structured logic of Operations Research (OR). Specifically, we introduce an \textit{OR-guided ``Pretrain-then-Reinforce"} paradigm. In this paradigm, reinforcement learning fine-tunes a pre-trained deep replenishment model, using reference solutions from a structured, multi-category OR model as the core guidance. This process directs the replenishment model toward a business-aligned decision logic, enhancing its generalization across diverse scenarios without incurring excessively high training costs.
\section{Models}
\label{sec:models}
\begin{figure}[h]
    \centering
    \caption{Overall Model Architecture}
    \includegraphics[width=0.95\linewidth]{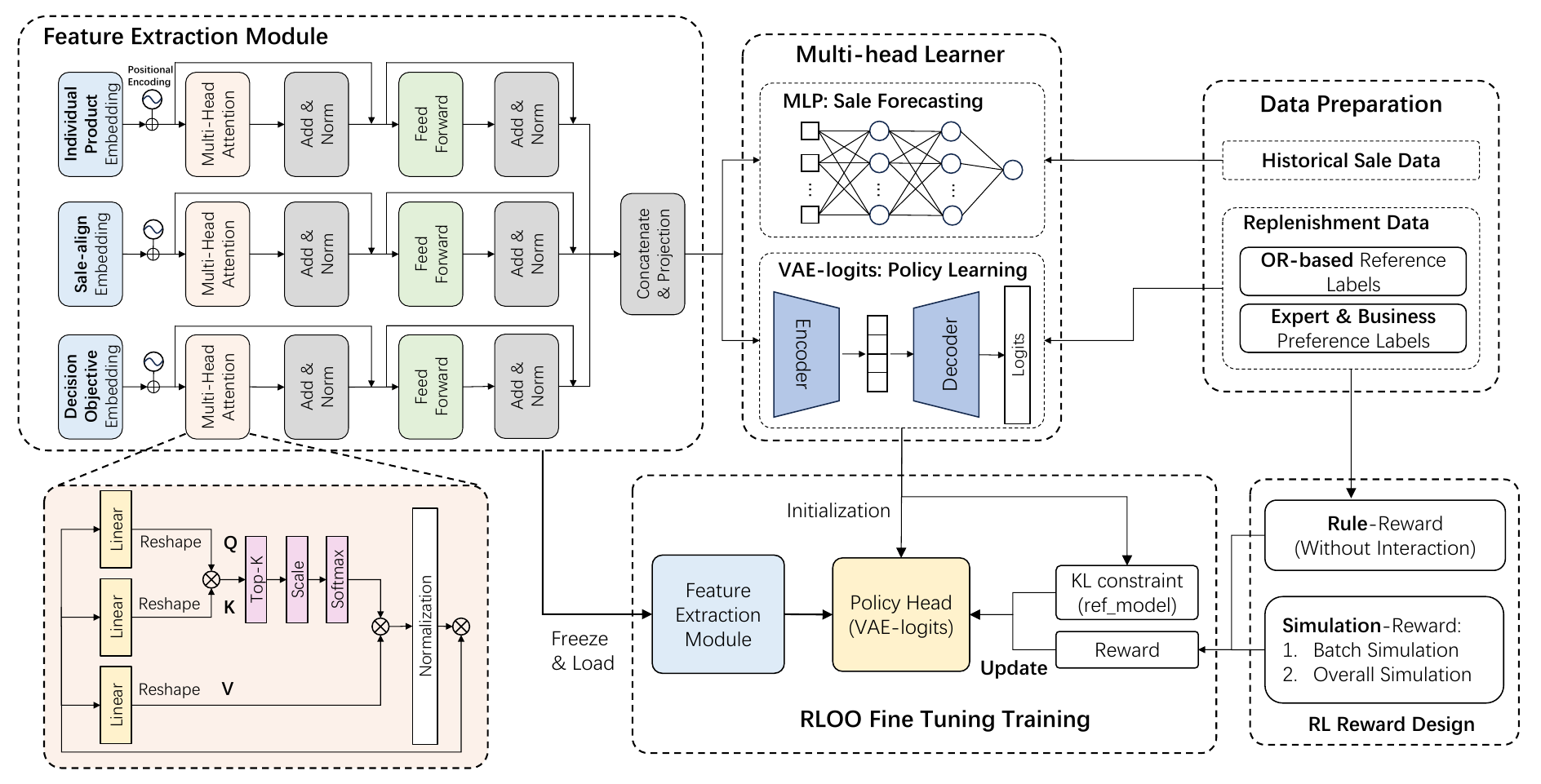}
    \label{fig:model_arch_full}
\end{figure}

This section details the architecture of our OR-Guided ``Pretrain-then-Reinforce" framework. We begin in Section~\ref{sec:DL_model} by presenting the deep learning (DL) model, whose design is informed by domain-specific insights for the pre-training stage. In Section~\ref{sec:RL_algorithm}, we detail the reinforcement learning (RL) algorithm for alignment and fine-tuning, discussing its advantages over conventional deep RL methods. Finally, and crucially, Section~\ref{sec:OR_model} presents the embedded inventory optimization model that provides the essential guidance for the training and alignment phases. These components constitute our integrated framework, which is built on a core philosophy: using the structured, model-based logic of OR to steer the powerful learning and alignment capabilities of AI. The overall architecture is illustrated in Figure~\ref{fig:model_arch_full}.

\subsection{Pre-trained Deep Neural Network Foundation for Replenishment}
\label{sec:DL_model}
The foundation model is designed to provide a strong baseline replenishment policy by learning a mapping function $f(\cdot)$ that directly outputs the SKU-level replenishment quantity $\hat{a}$ based on an observed feature vector $x$. These features include historical demands, past replenishment decisions, item-specific attributes, and temporal information. In the pre-training phase, we instantiate $f$ as a deep neural network and fit it to historical (feature-decision) data. Specifically, given a dataset of $N$ feature-optimal decision pairs $\{(x_i, a_i^*)\}_{i=1}^N$, where $a^*$ denotes the reference optimal decision (to be detailed in Section~\ref{sec:OR_model}), we learn the mapping by solving

$$
\min_{f : \mathcal{X} \to \mathbb{R}} \sum_{i=1}^N L\bigl(f(x_i),\,a_i^*\bigr)\,,
$$
where $L(\hat{a},a^*)$ is a suitable loss function measuring the deviation between the predicted action $\hat{a}_i=f(x_i)$ and the target $a_i^*$. Informed by domain expertise, we develop a Transformer-based architecture and a three-stage training strategy to ensure the model's efficient convergence and performance, which we detail in the following subsections.

\subsubsection{Neural Network Structure.} 
\paragraph{Feature Extraction Module.} The feature extraction module is designed to process three heterogeneous input streams for both sales forecasting and replenishment decision tasks:
\begin{itemize}[leftmargin=*, noitemsep] % noitemsep：减小列表行间距（学术更紧凑）
    \item Item-level sales data, encompassing historical sales time-series and initial forecasts of sales performance over the upcoming 90-day horizon;
    \item Item-specific inherent attributes, including product name, product category, and recent merchandise traffic metrics;
    \item Decision-objective-oriented features, specifically the product's sales contribution ratio within its respective category and human-expected turnover targets for the subsequent month.
\end{itemize}
Thereafter, a dedicated multi-layer Transformer encoder is deployed for each input stream individually. A dynamic weight-learning module is then leveraged to aggregate the distinct streams, with the resultant feature representations further fused into a unified feature embedding. This design enables the joint representation of static product attributes and temporal dynamics, providing a robust informational basis for the downstream decision module.

\paragraph{Multi‑Objective Learner.} The multi-objective learner then takes the unified feature embedding from the encoder as input and is trained on two distinct objectives: sales (demand) forecasting and replenishment decision-making. The \textbf{sales forecasting head}, also referred to as a calibration head, consists of a multi-layer perceptron (MLP) network. It generates point predictions for future demand, providing direct predictive feedback that monitors the decision head training. The \textbf{replenishment decision head}, which is the core component, is responsible for making the final decisions. We implement this using a VAE-logits framework:
$$
q_{\phi}(z \mid x_t) \;=\; \mathcal{N}\bigl(\mu_{\phi}(x_t),\,\mathrm{diag}(\sigma_{\phi}^2(x_t))\bigr), 
\quad
p_{\theta}(a \mid z) \;=\;\mathrm{Softmax}(W z + b),
$$
where $x_t$ denotes the fused input embedding at time $t$, and $z$ is a latent vector.
The variational posterior $q_\phi$ is defined by the predicted mean $\mu_\phi$ and variance $\sigma^2_\phi$. For the generative distribution $p_\theta$, the parameter $\theta=\{W,b\}$ serves as a linear projection layer, mapping the latent state $z$ to the logits of the replenishment action $a$.
This decision head is trained by maximizing the Evidence Lower Bound (ELBO). The VAE framework improves the generative capacity of the model. Importantly, the VAE-logits architecture outputs raw logits rather than the smoothed, probability-normalized results from a softmax layer. This design preserves the strength and heterogeneity of the signal, thereby facilitating exploration and convergence during downstream reinforcement training.

\subsubsection{Training Strategy.}
Our training strategy addresses two key challenges: the objective misalignment between forecasting and decision-making, and the scarcity of decision labels relative to abundant demand data. To navigate these nuances, we design a three-stage training strategy. First, we train the feature extraction module and the forecasting head on large-scale demand data. Next, with these modules frozen, we train the replenishment decision head on OR-derived labels (as detailed in Section~\ref{sec:OR_model}) to establish a baseline policy. Finally, we unfreeze all parameters and jointly fine-tune the entire model solely on the decision loss, while retaining the forecasting loss only for monitoring purposes.

This ``decouple-then-collaborate" strategy offers several advantages over the simpler approach that relies solely on decision data for training. First, by training the feature extraction and forecasting modules on large-scale sales data, the feature extractor learns robust and generalizable temporal representations. This significantly accelerates convergence when fitting the decision head with relatively scarce OR-based labels and substantially reduces the risk of overfitting. Second, using the auxiliary task of sales forecasting serves as a powerful form of regularization for the feature extractor, enhancing the stability and generalization of the final replenishment decisions.

\subsection{Reinforcement Learning Fine-Tuning with Feedback}
\label{sec:RL_algorithm}
While the pre-trained DL model generates fast and effective replenishment decisions, it suffers from two critical limitations: deviations from operational optimality and a lack of adaptive capability. Deviations arise from both the inherent fitting approximations of deep learning and the gap between theoretical OR solutions and complex real-world optima. Crucially, the model's inherently static nature struggles to accommodate non-stationary business dynamics, such as intermittent promotions, where the failure to adapt to shifting optimal policies can lead to substantial decision errors. To bridge this gap, we introduce Reinforcement Learning (RL) not merely as an optimizer, but as the critical \textit{alignment mechanism} to internalize OR decision patterns.

To implement this mechanism, we introduce an online reinforcement learning stage. This stage fine-tunes the pre-trained model using composite rule-based and simulation-based reward functions, to simultaneously align with OR decision patterns while exploring differentiated, SKU-level strategies for operational optimality. Our approach adapts the Reinforcement Leave-One-Out (RLOO) algorithm, originally developed to fine-tune large language models \citep{ahmadian2024back}. Selected for its high throughput and superior sample efficiency, RLOO is uniquely suited to our setting, offering the computational efficiency needed to learn from diverse, often scarce, alignment labels.

Central to this alignment process is the flexibility in defining these targets. Depending on the specific operational objective, reference labels can be sourced from historical business data to instill expert experience (e.g., accommodating promotional surges), or generated via the procedure in Section~\ref{sec:label_generating} to enforce adherence to structured OR decisions. Through the fine-tuning process against these diverse targets, the AI agent effectively internalizes the structural OR and business logic, ensuring the final policy is mathematically rigorous and operationally responsive.

\subsubsection{Reward Design.}
Our objective is to develop an RL algorithm that efficiently aligns with a reference policy while maintaining sufficient flexibility to explore and discover superior strategies. Rule‑based rewards directly encode the ideal OR‑optimal solution structure or business preferences, promoting stable and efficient convergence. In contrast, simulation-based rewards leverage interactive feedback from a replenishment simulator, exploiting RL’s trial-and-error learning capabilities, thereby facilitating decision improvement at the SKU level. To combine the strengths of both approaches, we define a hybrid reward function as
\[
    r_{total}=\omega \cdot r_{rule}+(1-\omega)\cdot r_{sim},
\]
where the hyperparameter $\omega$ balances rule-guided stability and simulation-driven adaptability.

The \textbf{rule-based reward} is computed solely from the input data, without interacting with the environment, ensuring high computational efficiency. We employ a sign-aware loss function to capture the economic asymmetry inherent in inventory problems (e.g., the cost of over-stocking versus under-stocking).  Let $\hat{\Delta}$ denote the prediction and $\Delta^{*}$ the reference decision provided by OR model or expert judgment. For an individual sample $i$ (i.e., one replenishment decision for a single SKU), we define the basic squared error $e^{(i)}=(\hat{\Delta}^{(i)}-\Delta^{*(i)})^2$. The sample-level loss is defined as
\[
    \ell^{(i)}=\bigl(1-\exp(-w_{\text{sign}}^{(i)}e^{(i)})\bigr)^{\gamma}\cdot w_{\text{sign}}^{(i)}e^{(i)},
\]
where $\gamma\ge 0$ controls the focal effect and the weight is defined as
\begin{equation*}
\label{eq:wsign}
w_{\mathrm{sign}}^{(i)} =
\begin{cases}
1, & \text{if }\operatorname{sign}\bigl(\hat{\Delta}^{(i)}\bigr)
= \operatorname{sign}\bigl(\Delta^{*(i)}\bigr),\\
\alpha, & \text{if }\operatorname{sign}\bigl(\hat{\Delta}^{(i)}\bigr)
\neq \operatorname{sign}\bigl(\Delta^{*(i)}\bigr),
\end{cases}
\quad \alpha>1.
\end{equation*}

When integrating this rule-based loss into a reward framework for policy learning, we take the rule-reward to be the negative loss at the batch level: $r_{rule}=-\frac{1}{|\mathcal{B}|}\sum_{i\in\mathcal{B}}\ell^{(i)}$.

By contrast, the \textbf{simulation‑based reward} relies on interacting with a replenishment simulator and is designed to drive overall multi‑objective improvement. Specifically, for a given state $s$ and action $a$, we interact with the simulator to project the resulting inventory trajectory over a future horizon of  $H$ days, evaluating key performance indicators such as inventory turnover $T$ and lost sales $L$. We define a Pareto-dominance indicator
\begin{equation*}
    r_{sim}(s,a)=
    \begin{cases}
        +1,\ (T,L)\succ (T_0,L_0)\\
        -1,\ (T,L)\prec (T_0,L_0)\\
        0,\ \text{otherwise}
    \end{cases}
\end{equation*}
where $(T_0,L_0)$ are the baseline metrics achieved by the reference action $a^*$. This discrete reward scheme directly incentivizes the agent to explore policies that Pareto-dominate the reference decision. The batch-level reward is $r_{sim}=\frac{1}{|\mathcal{B}|}\sum_{s\in\mathcal{B}}r_{sim}(s,a)$.

\subsubsection{Policy Optimization.}
We adopt the \textbf{Reinforcement Leave‑One‑Out (RLOO)} algorithm to fine-tune the policy. RLOO offers several advantages over common alternatives. Unlike PPO, RLOO requires neither a critic nor a learned reward model, which reduces training variance. Compared with DPO, its online RL design enables exploration beyond pure alignment, leading to higher sample efficiency and faster convergence. These properties are also particularly valuable for efficiently fine-tuning our replenishment model. In practice, RLOO minimizes the following objective function:

$$
J_{\mathrm{RLOO}}(\theta) \;=\; \mathbb{E}_{x\sim\mathcal{D}}\;\mathbb{E}_{y^{(1)},\dots,y^{(k)}\sim\pi_\theta(\cdot\mid x)}
\left[\frac{1}{k}\sum_{i=1}^k \bigl(R^{(i)} - b^{(i)}\bigr)\,\log\pi_\theta\bigl(y^{(i)}\mid x\bigr)\right],
$$
where
$$
R^{(i)} = r_{total}(x,y^{(i)}) \;-\;\beta\,\log\frac{\pi_\theta(y^{(i)}\mid x)}{\pi_{\mathrm{ref}}(y^{(i)}\mid x)}
$$
combines the reward $r_{total}$ with a KL-divergence penalty term (with balance coefficient $\beta>0$), and $b^{(i)}$ is the leave‑one‑out baseline

$$
b^{(i)} = \frac{1}{k-1}\sum_{j\neq i} R^{(j)},
$$
which reduces variance without introducing bias. This formulation effectively calibrates and improves our replenishment policy, while preventing excessive deviation from the initial pre‑trained policy. Algorithm~\ref{alg:RL_alignment} presents the full RLOO fine-tuning procedure.

\begin{algorithm}[hbt]
\caption{RLOO Algorithm for Alignment and Fine-Tuning}
\label{alg:RL_alignment}
\begin{algorithmic}[1]
\Require Pre-trained policy $\pi_{\theta_0}$; reference policy $\pi_{\text{ref}}$; dataset $\mathcal{D}$ of decision traces; simulator $\mathcal{S}$; rule-based reward weight $\omega \in [0,1]$; KL penalty coefficient $\beta$; batch size $B$; sample length $k$.
\State Initialize $\theta \gets \theta_0$ from pre-trained DL model.
\For{each batch $\{(x_i, a_i^*)\}_{i=1}^B \subset \mathcal{D}$}
    \For{$i = 1, \dots, B$}
        \State Sample $k$ actions $\{y^{(j)} \sim \pi_\theta(\cdot \mid x_i)\}_{j=1}^k$.
        \For{$j = 1, \dots, k$}
            \State \textcolor{blue}{\textit{Rule-based Reward:}}
            \State Compute adjustment error $e^{(j)} = (y^{(j)} - a_i^*)^2$ and sign weight $w_{\text{sign}}^{(j)}$.
            \State Compute focal loss $\ell^{(j)} = (1 - \exp(-w_{\text{sign}}^{(j)} e^{(j)}))^\gamma \cdot w_{\text{sign}}^{(j)} e^{(j)}$ and set $r_{\text{rule}}^{(j)} = -\ell^{(j)}$.
            \State \textcolor{blue}{\textit{Simulation-based Reward:}}
            \State Execute $y^{(j)}, a_i^*$ in simulator $\mathcal{S}$ to get metrics $(T^{(j)}, L^{(j)})$ and $(T_0, L_0)$, set $r_{\text{sim}}^{(j)}$.
            \State \textcolor{blue}{\textit{Hybrid Reward:}}
            \State Total reward $r_{\text{total}}^{(j)} = \omega \cdot r_{\text{rule}}^{(j)} + (1-\omega) \cdot r_{\text{sim}}^{(j)}$, and $R^{(j)} = r_{\text{total}}^{(j)} - \beta \log \frac{\pi_\theta(y^{(j)} \mid x_i)}{\pi_{\text{ref}}(y^{(j)} \mid x_i)}$.
        \EndFor
        \For{$j = 1, \dots, k$}
            \State Compute leave-one-out baseline $b^{(j)} = \frac{1}{k-1} \sum_{\ell \neq j} R^{(\ell)}$.
        \EndFor
    \EndFor
    \State Update $\theta$ via gradient descent on $\mathcal{J}_{\text{RLOO}}(\theta) = \frac{1}{B} \sum_{i=1}^B \frac{1}{k} \sum_{j=1}^k (R^{(j)} - b^{(j)}) \log \pi_\theta(y^{(j)} \mid x_i)$.
\EndFor
\end{algorithmic}
\end{algorithm}

\subsection{An Embedded Simulation-Augmented Inventory Optimization Model}
\label{sec:OR_model}
The OR model is the analytical core of our framework, designed with a specific purpose: to provide structured, operationally-grounded guidance for the AI Agent. Its architecture is intentionally concise and augmented by a high-fidelity simulation, a deliberate design choice that enables a clear division of labor between the AI and OR components. The contribution of our OR model to the framework lies not in its explicit structural complexity, but in its ability to generate high-quality reference decisions by integrating the analytical rigor of optimization with rich insights from the simulation of the operational environment.

\subsubsection{Model Formulation.}
We consider a single-period, multi-category joint inventory optimization problem. We define the decision variable $X_i$ as the target replenishment quantity for category $i$, measured in \textbf{inventory days} and restricted to an integer value $v\in [L,U]$ to align with JD.com's key performance indicators. Operationally, this corresponds to the target number of days of demand the replenishment quantity should cover. For example, a decision of $X_i=7$ for category $i$ indicates that the quantity is sufficient to meet one week of expected demand. This choice, as opposed to the traditional use of raw replenishment quantities, unifies the decision unit across heterogeneous products (e.g., a 30-day supply of salt vs. a 30-day supply of refrigerators), substantially reduces the decision space for computational tractability, and aligns directly with industrial key performance indicators such as inventory turnover days. We introduce a binary indicator variable $X_{i,v}$ to indicate whether category $i$ chooses to replenish for $v$ days, then the replenishment decision for category $i$ can be expressed as 
\begin{equation*}
    X_i \;=\;\sum_{v=L}^{U} v\cdot X_{i,v}.
\end{equation*}

The optimization problem involves two competing objectives: minimizing inventory levels while reducing lost sales, where inventory is quantified by its value to align with the monetary unit of sales losses. Instead of using a weighted linear combination of the two conflicting objectives, we address this practical trade-off by reframing the lost sales objective as a hard constraint.
Let $stock_{i,v}$ denote the value of the cumulative inventory for category $i$ over the planning horizon, given a target of $v$ inventory days, and let $loss_{i,v}$ be the lost sales for category $i$ under the replenishment quantity $v$ inventory days. The complete optimization problem is as follows:
\vspace{-5pt}
\begin{align}
\min_{\,X_{i,v},X_i}\quad & \sum_{i=1}^I\sum_{v=L}^U X_{i,v}\cdot stock_{i,v} \nonumber\\
\text{s.t.}\quad 
& \sum_{v=L}^U X_{i,v} = 1, 
\quad \forall i=1,\dots,I, \label{eq:con_single_selection}\\
& \sum_{i=1}^I\sum_{v=L}^U X_{i,v}\cdot loss_{i,v}
\;\le\; 
SALE\,\bigl(1 - \alpha_{\text{loss}}\bigr), \label{eq:con_lost_sales}\\
& X_{i,v} \in \{0,1\},\quad\forall i=1,...,I, v\in \{L,..,U\}. \label{eq:con_binary}
\end{align}

Here, $SALE$ represents the total sales realized during the planning period. The objective function minimizes the cumulative inventory value across all categories. Constraint \eqref{eq:con_single_selection} enforces that exactly one replenishment quantity is chosen for each category. Constraint \eqref{eq:con_lost_sales} reformulates the lost sales minimization objective into a hard constraint governed by a hyperparameter $\alpha_{\text{loss}}$, which represents the maximum allowable stockout rate. A key advantage of this reformulation is that varying $\alpha_{\text{loss}}$ facilitates the generation of a set of Pareto-optimal solutions. This capability is instrumental in producing a rich set of candidate labels for AI training, as detailed in our labeling approach in Section~\ref{sec:label_generating}.

A critical feature of our model is the way we calibrate the values of key parameters, $stock_{i,v}$ and $loss_{i,v}$. Instead of relying on simplified cost assumptions and explicit modeling of complex cross-category demand correlations (such as substitution effects), we employ a streamlined, \textbf{trace-driven simulation} process. This approach prioritizes empirical fidelity over structural complexity. Unfolding over a defined horizon $T$ and taking as input historical demand realizations and real-world constraints (e.g., Nominal Review Time, Vendor Lead Time, and specific cost structures), the simulator iteratively ``replays" the operational timeline for every candidate replenishment quantity $v$—updating inventory states based on order arrivals, demand fulfillment, and lead-time delays to aggregate the cumulative inventory value and sales loss. By doing so, it implicitly captures intricate market dynamics without the risk of model misspecification. This strategic design enables the rapid, exhaustive counterfactual evaluation of all candidate decisions to determine their realized performance under realistic conditions. The derived metrics then serve as the precise coefficients for our optimization model. Algorithm~\ref{alg:simulation_process} outlines this procedure. Section~\ref{sec:OR_labels_valid} validates the effectiveness of these OR-derived labels, demonstrating their ability for accurate and proactive stock preparation aligned with sales rhythms.

\begin{algorithm}[h]
\caption{Simulation Logic for Parameter Generation}
\label{alg:simulation_process}
\begin{algorithmic}[1]
\Require 
    Set of SKUs $\mathcal{I}$; Candidate inventory days $\mathcal{V}$; Horizon $T$; 
    Demand data $\mathbf{D}$; 
    Parameters: Review Period ($NRT$), Lead Time ($VLT$), Unit Cost ($c$), Unit Price ($p$).
\Ensure 
    Inventory value $stock_{i,v}$ and sales loss $loss_{i,v}$.
\For{each SKU $i \in \mathcal{I}$ and decision $v \in \mathcal{V}$}
    \State Initialize state variables ($I_0, \text{Pipeline}_0$) and metrics ($stock_{i,v}, loss_{i,v} \gets 0$).
    \State Use recent average daily demand $\bar{d}_i$ to map inventory days $v$ to units $Q_{i,v}\approx v\times \bar{d}_i$.
    \For{$t = 1 \to T$}
        \State \textbf{Inventory Update:} $I_t \gets I_{t-1} + \text{Arrivals}_t$. %\Comment{Order arrival}
        \State \textbf{Demand Realization:}\quad $\text{Lost}_t \gets (d_{i,t} - I_t)^+$; \ $I_t \gets (I_t - d_{i,t})^+$. %\Comment{Lost sales logic}
        \State \textbf{Metric Accumulation:} \quad $stock_{i,v} \mathrel{+}= c_i \cdot I_t$; \ $loss_{i,v} \mathrel{+}= p_i \cdot \text{Lost}_t$.
        \State \textbf{Replenishment Review:}
        \If{$t$ is the replenishment ordering day}
            \State Order $q_t \gets Q_{i,v}$. Schedule arrival of $q_t$ at $t + VLT_i$.
        \EndIf
    \EndFor
\EndFor
\end{algorithmic}
\end{algorithm}

\subsubsection{Design Philosophy and Rationale.}
Our OR model departs in meaningful ways from traditional inventory formulations, reflecting both the real operational challenges observed in our collaboration with JD.com and the guiding principles underlying our AI+OR framework. We justify our approach based on two core principles.

\textit{First, structural conciseness facilitates a clear division of labor.} As we have argued, OR model's power comes from its function as a guidance generator, which integrates analytical optimization with rich simulation insights, rather than from its own structural complexity. Embedding complex, multi-period dynamics and explicit cost structures directly into the OR model would be counterproductive. It would risk creating confounding signals, blurring the line between the explicit logic of the OR model and the implicit patterns learned by the AI, which could degrade the quality of the training labels. By keeping the OR model structurally concise, we ensure it provides a clear, unambiguous signal about the optimal economic trade-offs, serving as a stable ``compass" for the AI without dictating the complex path to get there.

\textit{Second, the model does not ignore complexity, but rather embeds it through simulation.} Specifically, our model is simulation-augmented: its key parameters, $stock_{i,v}$ and $loss_{i,v}$ are not static assumptions. Instead, they are the output of a high-fidelity simulator that runs on historical data, incorporating actual SKU-level costs, complex business constraints, and observed demand patterns. This approach transcends the limitations of explicit modeling. It allows us to endogenously account for real-world dynamics, inter-category correlations, and subtle operational nuances that are intractable to formulate mathematically. By outsourcing this complexity to the simulator, the OR model can focus on optimizing the core trade-offs, yielding reference decisions that are both analytically sound and deeply grounded in operational reality.

In concert, these design principles allow our OR model to produce high-quality, differentiated reference decisions that effectively guide the AI, bridging the gap between tractable optimization and the immense complexity of large-scale e-commerce inventory management.

\subsubsection{Label Generation.} 
\label{sec:label_generating}
The final step in leveraging our OR model is to generate high-quality and operationally-sound labels for training the AI components. Since our OR formulation produces a set of Pareto-optimal solutions by tuning the hyperparameter $\alpha_{\text{loss}}$, we face a critical challenge: selecting the single solution that best reflects the true (and often implicit) business preferences. Conventional hyperparameter tuning methods like cross-validation are ill-suited for this task. Such methods are inherently preference-agnostic; they would select an $\alpha$ value based on a purely numerical metric, potentially yielding solutions that are mathematically optimal but operationally infeasible. For instance, a numerically optimal solution might drastically reduce inventory turnover in certain categories, which is an unacceptable outcome in real operations.

To overcome this limitation, we introduce a business-metric calibration procedure. Specifically, we perform a binary search on $\alpha_{\text{loss}}\in[0,1]$, using the degree of \textit{alignment with real-world inventory turnover} as the criterion. For each candidate $\alpha_{\text{loss}}$, we solve the OR model and simulate the resulting policy on historical demand to assess the achieved turnover. The final training labels correspond to the replenishment decisions that best reproduce the real turnover pattern.

Although this calibration does not guarantee numerical optimality, it provides an operationally sound mechanism that yields reference decisions capturing a managerially consistent trade-off between inventory holding and service level. This systematic, data-driven procedure aligns the theoretical solutions of OR with the complex and often uncodified preference of real-world business practice. A key advantage is that it requires no manual specification of objective weights, making it particularly valuable in practical settings where priori knowledge is typically unavailable.

\subsection{Model Scale and Discussion}
The resulting model is computationally lightweight, with approximately \textbf{0.93 million} parameters in total. Under a typical configuration, the majority of parameters are concentrated in the VAE component and the Transformer encoder, which together account for approximately 78\% of the total and are responsible for the model's core capabilities in feature extraction and representation. The remaining 22\% of parameters constitute the reinforcement learning policy network, which enables adaptive decision-making during the fine-tuning stage. This compact design enables efficient training and deployment in resource-constrained environments, while the OR-guided pre-training and fine-tuning paradigm ensures adaptability and accuracy for practical applications.

% 需要详细讨论，并且注意表述；AI+OR, 如何分工。
% 首先，不为通用能力开发而是专注业界实际问题解决设计模型，这种聚焦确实可以较大降低所需规模；
% 但是我们认为以下两点至关重要，并揭示了关键见解。
% 1. 我们的“预训练+微调”框架和训练策略降低了对高质量决策数据的需求，而我们domain experties informed模型架构避免了不必要的复杂组件。
% 2. but may be the most important, 我们的AI和OR模型相互配合。AI通过复杂结构提供识别复杂决策规律的能力，而通过仿真程序捕捉真实动态性和相关性的OR model，尽管形式简洁，但提供了关键的结构化和analytical的guiding。这种配合呼应了我们对于AI+OR的基本思路，最终使得模型输出能与最佳决策背后的逻辑保持一致，从而在没有大规模模型结构或大规模决策数据的情况下实现强大的结果。
This compact model size stands in stark contrast to the scaling laws that dominate large language model development, a phenomenon we term ``anti-scaling." While focusing on the specific problem of inventory management rather than general intelligence certainly reduces complexity, we argue that two additional factors are crucial and reveal deeper insights into the synergy of AI and OR.

\textit{First, our methodological design promotes efficiency.} The ``pretrain-then-reinforce" architecture provides a cohesive structural basis for synergizing AI capabilities with OR logic. Our three-stage training strategy strategically leverage abundant demand data, significantly reducing the reliance on scarce, high-quality decision labels. Furthermore, our domain-informed model design consciously avoids the unnecessary complexity and scale of general-purpose foundation models.

\textit{Second, and most critically, our framework is built upon a synergistic division of labor between AI and OR.} The AI component, with its complex neural architecture, excels at perception—discerning intricate patterns from high-dimensional features. The OR model, though structurally concise, provides the essential structured guidance. Augmented by a simulation that captures real-world dynamics, it offers analytically-grounded and sound reference decisions. This partnership, which embodies the core philosophy of our work, aligns the model’s outputs with the underlying logic of optimal decisions, enabling it to achieve powerful results without the need for a massive model or vast quantities of decision data.
\section{Numerical Experiment}
\label{sec:num_exp}
Our collaboration with JD.com on this project was initiated in January 2025. JD.com is one of the largest online retailers in China, owning more than 1,500 warehouses to manage inventory and has replenishment agreements with tens of thousands of vendors. To ensure the reliability of our model before live deployment, we conducted a rigorous, multi-phase validation process from June to August 2025. In this section, we detail this offline evaluation phase based on large-scale historical real-world data from JD.com. We begin in Section~\ref{sec:OR_labels_valid} by examining the effectiveness of the decisions generated by our simulation-augmented OR model, highlighting their high degree of consistency with real-world business logic. Subsequently, Section~\ref{sec:comparisons} benchmarks our framework against established baselines, demonstrating its significant performance advantages. Finally, Section~\ref{sec:value_RL} validates the RL fine-tuning mechanism, using a real-world promotional case study to demonstrate its efficacy in aligning decisions with dynamic business requirements.

\subsection{Validation of OR-based Labels and Analysis of Differentiated Policies}
\label{sec:OR_labels_valid}
In this section, we validate our labeling method's ability to generate differentiated replenishment decisions for a large-scale, heterogeneous set of SKUs that effectively capture key business preferences. Specifically, we utilized our OR model to generate ``optimal" reference replenishment decisions for a series of product categories, grounded in real historical data. These decisions are represented in terms of ``inventory days," serving as labels used in training phases.

\paragraph{Decision Heterogeneity.} To illustrate the differentiated nature of OR-based replenishment decisions across product categories, we examine the decisions generated by our OR model under the classical ABC–XYZ classification, which jointly accounts for product value (A–C, from highest to lowest) and demand stability (X–Z, from most to least stable). We acknowledge that alternative classification methods could also be adopted in practice. 
\vspace{-8pt}
\begin{figure}[h]
    \centering
    \caption{OR-derived replenishment quantity by product categories (2024)}
    \includegraphics[width=0.8\linewidth]{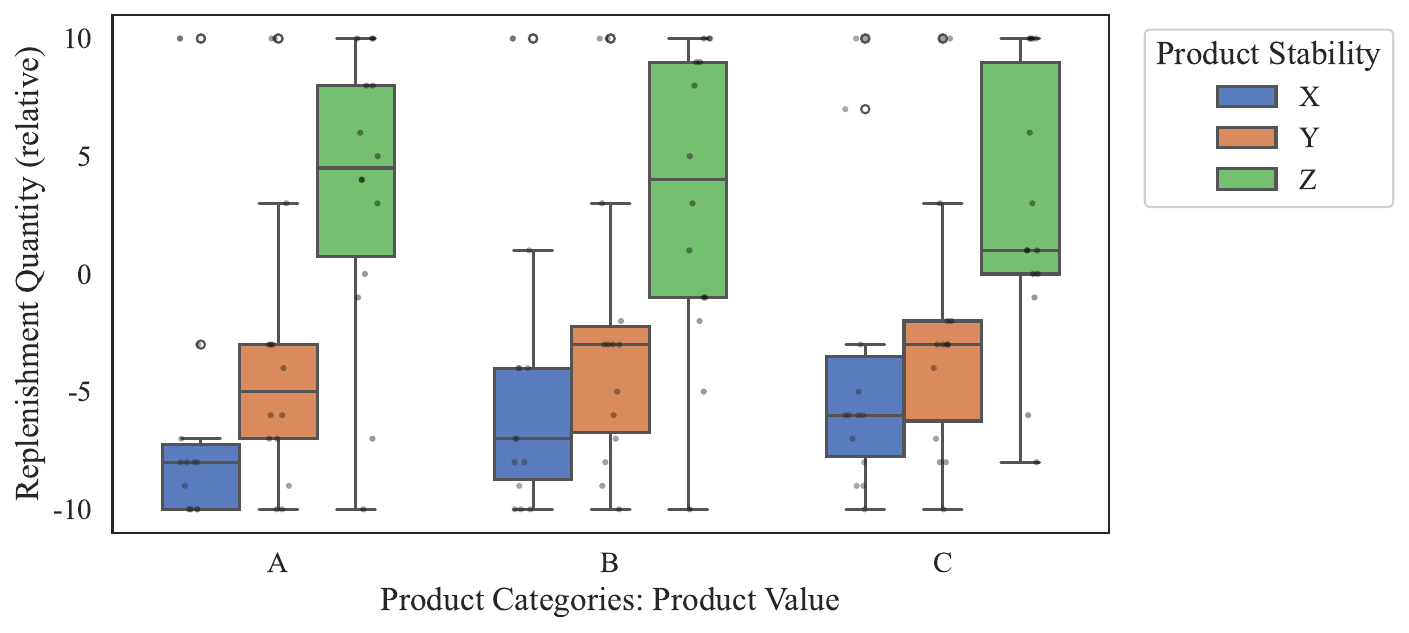}
    \label{fig:or-label distribution}
\end{figure}
\vspace{-18pt}
As shown in Figure~\ref{fig:or-label distribution}, the OR model produces systematically varied replenishment strategies across product types. Categories with high turnover and stable demand tend to receive lower replenishment recommendations to enhance capital efficiency. Conversely,  products with slower turnover or higher demand volatility are assigned higher quantity, reflecting the need to buffer against uncertainty. Overall, these results demonstrate that the OR model endogenously captures business-consistent heterogeneity, aligning quantitative decisions with intuitive operational logic.

\begin{figure}[h]
    \centering
    \caption{Temporal consistency and reusability of OR-derived replenishment decisions}
    \includegraphics[width=0.8\linewidth]{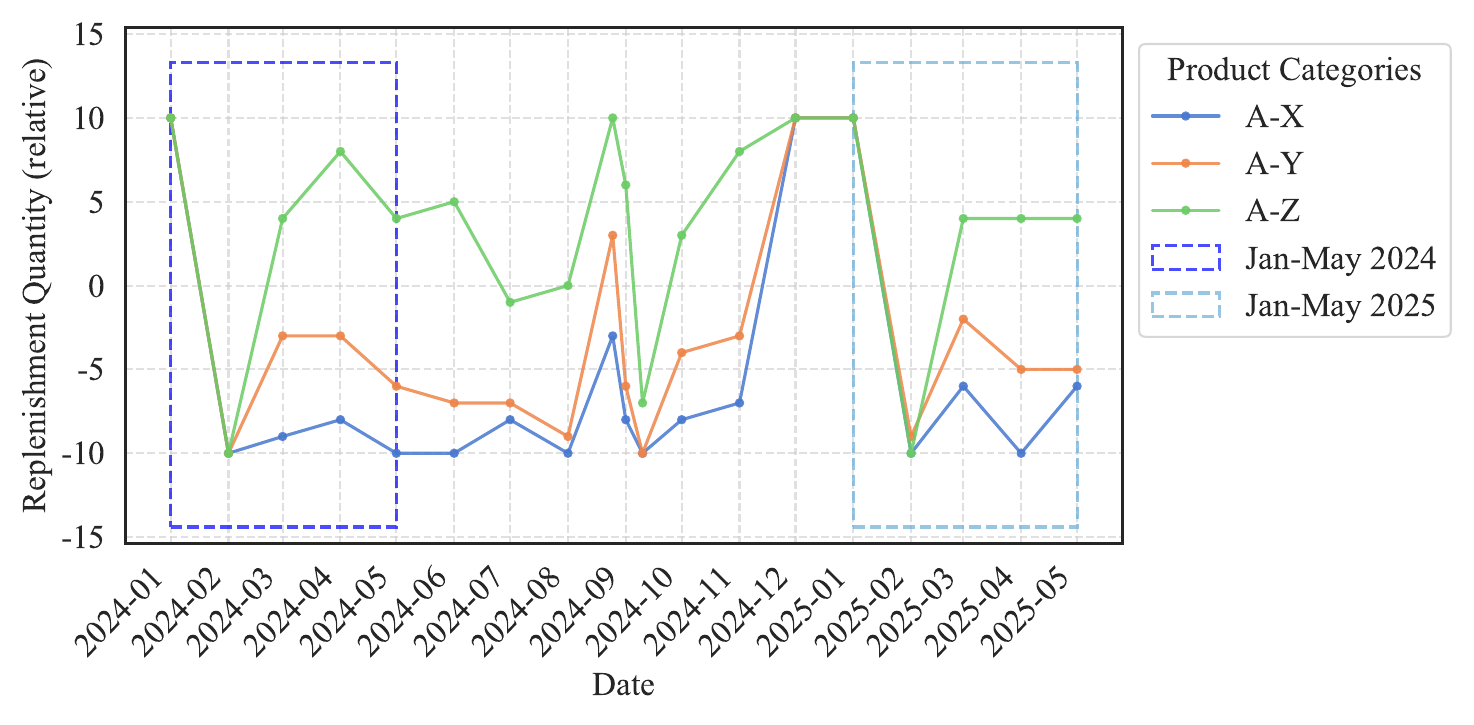}
    \label{fig:or-label trend}
\end{figure}

\begin{figure}[h]
    \centering
    \caption{Alignment of OR-derived decision pattern with e-commerce promotion rhythm}
    \includegraphics[width=0.8\linewidth]{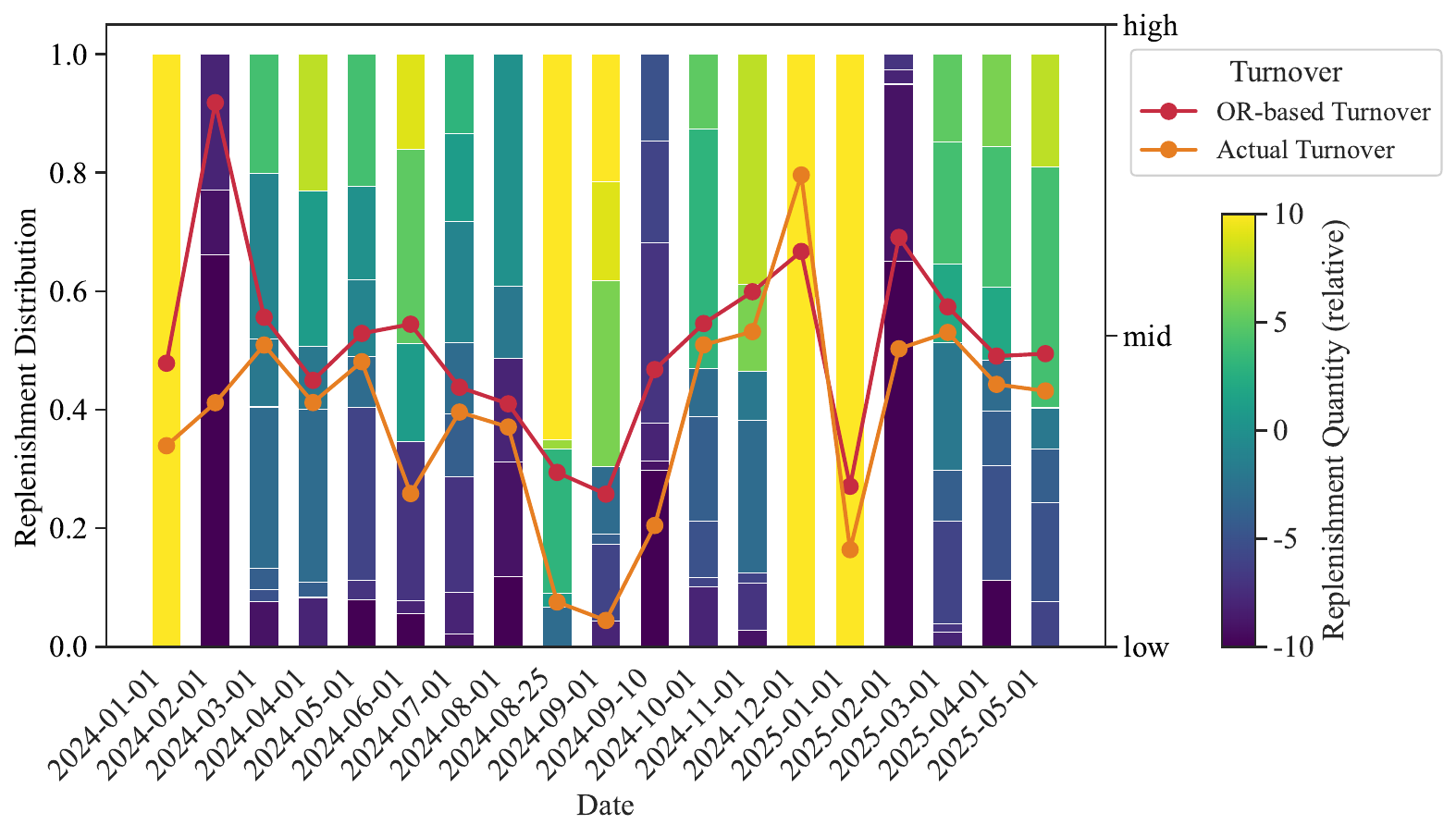}
    \label{fig:or-label distribution2}
\end{figure}

\paragraph{Transferability and Reusability of Differentiated Decisions.} To examine the transferability of OR-derived decisions, we compute the OR-optimal inventory targets for high-value (A) products across varying demand volatility levels from January 2024 to May 2025, as shown in Figure~\ref{fig:or-label trend}. While the absolute inventory levels differ across product categories, their temporal trajectories display notable consistency and a clear annual cyclicality. This pattern indicates that the OR model effectively captures recurring market dynamics such as seasonality and promotional cycles in a stable and interpretable manner. Such temporal coherence validates that the OR-based differentiated decisions are not only operationally sound but also transferable and reproducible across time. This property enables the deep model to learn generalizable decision logic from these labels.
\vspace{-10pt}
\paragraph{High sensitivity to real-world promotion rhythm.} Figure~\ref{fig:or-label distribution2} visualizes the temporal evolution of replenishment strategies. The stacked bars represent the distribution of replenishment decisions (weighted by sales contribution), where lighter shades denote higher replenishment quantities and darker shades indicate conservative decisions. The superimposed line plot tracks the turnover.

The results demonstrate that OR-based labels closely align with real-world promotional rhythms, exhibiting proactive and timely stock preparation. We observe significant inventory build-ups prior to major promotions such as the Chinese Spring Festival and China's Double 11 (the world's largest e-commerce shopping festival). The September 9th promotion serves as an intuitive example. As illustrated by the specific timeline from August 25 to September 10, the model initiates aggressive replenishment on August 25 (predominantly light-colored segments), precisely aligning with the typical 12-14 day stock preparation lead time. As the event approaches (September 1), the replenishment intensity tapers, shifting sharply to a conservative ``destocking" mode on September 10 (dominated by dark segments) immediately post-promotion. This observed ``pre-load and taper" pattern validates our OR model's ability to proactively manage lead times and prevent post-event overstocking, mirroring rational operational logic.

\subsection{Comparison with Various Replenishment Methods}
\label{sec:comparisons}
\subsubsection{Benchmarks.} To evaluate the performance of our proposed model, we compare it against a comprehensive set of benchmarks. These include several common parameterized predictive-then-optimize (PTO) methods, a pure end-to-end deep learning model, and JD.com’s current production-level replenishment system.

We first consider classic parameterized replenishment policies under the PTO framework. Under the assumption of i.i.d. normal demand, the base stock level can be calculated as follows:
$$
    \text{PTO}_{normal}=\mu_D\cdot(R+LT)+\phi^{-1}(\frac{b}{b+h})\cdot \sigma_D\cdot\sqrt{R+LT},
$$
where $R$ is the review period, $LT$ is the lead time, and $b$ and $h$ represent the stockout cost and holding cost respectively. The parameters $(\mu_D,\sigma_D)$ are estimated from historical demand data. Similarly, under the assumption that daily demand follows a Gamma distribution $Gamma(k,\theta)$, the total demand over the risk period $D_L$ follows $Gamma((R+LT)\cdot k,\theta)$. The corresponding base stock level is then:
$$
\text{PTO}_{gamma}=Q_L^{gamma}(\frac{b}{b+h}),
$$
where $Q_L$ is the quantile function for the total demand over $L$ days. 

Additionally, we include JD.com’s incumbent algorithm as a critical benchmark. This system employs a periodic review model, deriving decisions through optimization constrained by category-specific business objectives, such as in-stock rate and turnover targets. Notably, key parameters for this optimization are informed by pre-trained predictive models. Inclusion of this state-of-the-art industrial solution, alongside classic theoretical baselines, ensures a rigorous and comprehensive evaluation of our framework's practical efficacy.

\subsubsection{Results.}
We collectes and processes data within the top-level product category of ``snack foods," including 5,319 SKUs and 70,210 SKU-DC pairs from JD.com, to evaluates the performance of various methods using this real historical dataset. The evaluation metrics include \textit{In-stock Rate, Inventory Turnover Days, Holding Cost, Stockout Cost and the Total Cost}. The first two and total cost are of primary practical concern and represent our key optimization objectives. Specifically, the In-stock Rate is defined as the percentage of days without stockouts, while Turnover Days is calculated as the ratio of average inventory level to average daily demand. Both holding cost and stockout cost are precisely quantified using SKU-specific parameters. % We randomly partitioned our data into a training set and a validation set. For our model, we further partitioned the training set into a pre-training dataset for the deep learning (DL) model and a separate dataset for reinforcement learning (RL) fine-tuning.

\begin{table}[h]
    \centering
    %\small
    \caption{Comparison between different replenishment models}
    \label{tab:method_comparison}
    % 表格环境前添加\rowcolors开启行着色（仅对指定行生效）
    \begin{tabular}{lccccc}
    \toprule  % 顶部粗线
    Method        & Turnover Days & In-stock Rate & Holding Cost & Stockout Cost & Total Cost   \\
    \midrule  % 中间细线
    OR            & X        & Y             & 1287         & 1713          & 3000         \\
    PTO\_1        & X-1.85   & Y-1.79\%      & 1156         & 2226          & 3382(+12\%)  \\
    PTO\_2        & X-1.33   & Y-1.19\%      & 1193         & 2055          & 3248(+8\%)   \\
    BM\_50        & X-3.28   & Y-13.21\%     & 940          & 5505          & 6444(+114\%) \\
    BM\_85        & X+1.82   & Y-7.12\%      & 1289         & 3756          & 5045(+68\%)  \\
    JD online     & X+0.37   & Y-1.72\%      & 1285         & 2206          & 3491(+16\%)  \\
    \bfseries our model & \bfseries X-1.36 & \bfseries Y-0.85\% & \bfseries 1196 & \bfseries 1958 & \bfseries 3153(+5\%) \\  % 最后一行加粗
    \bottomrule  % 底部粗线
    \end{tabular}
\end{table}

Our experimental results are summarized in Table~\ref{tab:method_comparison}, where the cost values reported are desensitized results based on real outcomes. ``PTO1" and ``PTO2" represent the data-driven basic methods under the assumptions of a normal and gamma demand distribution respectively. ``$\text{BM}_{x}$" refer to the basic policy using $x$-th quantile prediction of demand as replenishment decision. ``OR" refers to the replenishment decisions derived from our OR model with respect to the real turnover, as described in Section~\ref{sec:OR_model}, which also serve as the reference labels for our DL and RL training. ``JD Online" represents JD.com's incumbent replenishment algorithm. To ensure a fair comparison, all benchmarks utilize the same demand forecasting module as our proposed framework.

The results demonstrate the superior performance of our proposed OR-Guided ``Pretrain-then-Reinforce" model across all metrics. Notably, our method achieves an optimal trade-off between inventory turnover days and in-stock rate, yielding the lowest total inventory cost among all evaluated methods. Furthermore, our approach exhibits comprehensive superiority over JD.com's incumbent algorithm: it reduced inventory turnover days by \textbf{1.73} and increased the in-stock rate by \textbf{0.87\%}. This operational efficiency drives a \textbf{9.68\%} reduction in total inventory costs (comprising a 6.96\% drop in holding costs and an 11.26\% cut in stockout losses). These results underscore that our framework, guided by optimality principles, delivers robust and state-of-the-art performance.

\subsection{The Value of Reinforcement Learning}
\label{sec:value_RL}
\subsubsection{Fast and Effective Alignment with Reference Decisions.}
\begin{figure}[h]
    \centering
    \caption{RL can help approximate human preference}
    \includegraphics[width=0.9\linewidth]{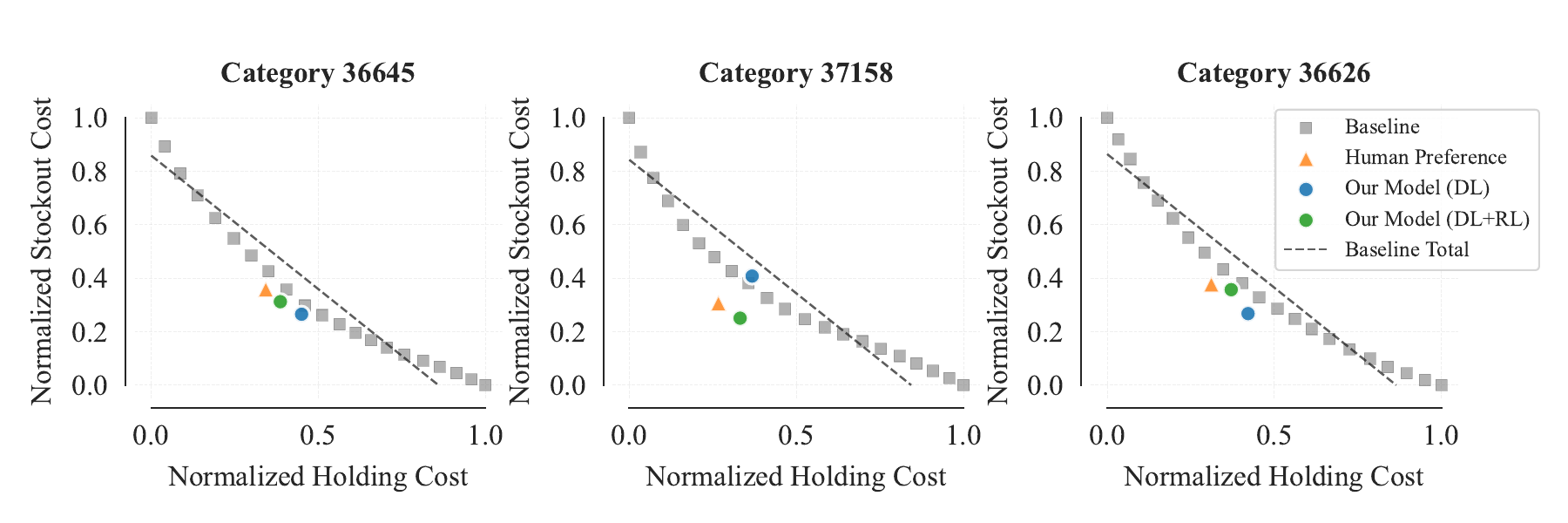}
    \label{fig:rl_value}
\end{figure}

Beyond alignment, a central motivation for RL fine-tuning is to exploit RL’s ability to learn replenishment policies that more closely approach the true operational optimum, thereby overcoming the inherent limitations of the simulation-augmented OR model. Consequently, we conducte separate tests within each secondary category of the ``snack foods" category. We compare the performance of several models, including the OR-optimal decisions, a pure deep learning model, and our RL fine-tuned deep learning model. The results for three representative categories are presented in Figure~\ref{fig:rl_value}. The figure clearly illustrates the quality improvements that are attributable to RL fine-tuning, relative to the end-to-end DL model's output: primarily, it pulls the model's output into close alignment with the rigorous OR benchmarks to correct inherent biases; simultaneously, it utilizes exploration to further refine these decisions, enabling superior trade-offs between the dual objectives.

\subsubsection{Case study: Adaptive Alignment RL Fine-Tuning for a Major Promotion.}
To demonstrate the adaptability and practical efficacy of our RL alignment mechanism, we conducte a real-world case study using data from a major promotional event at JD.com. Such events are characterized by volatile demand surges, necessitating proactive stock preparation to maintain high in-stock rates and mitigate stockout-induced losses.

Though the OR-guided RL model demonstrates robust performance in routine operations, it may exhibit undesirable conservatism during high-stakes periods such as major promotions. This limitation arises because the OR guidance targets general optimality across the full time horizon. Since routine periods dominate the data distribution, the optimization objective tends to ``regress towards the mean", prioritizing standard cost-efficiency trade-offs. Consequently, in infrequent but critical scenarios that require aggressive replenishment, the ``rational" OR guidance may inadvertently suppress necessary stock preparation. To solve this tension, our framework introduces a secondary fine-tuning step using a small set of expert decision data. This process enables the model to internalize scenario-specific imperatives that extend beyond the scope of general OR modeling, such as the implicit strategic mandate to maximize fill rates during promotions.

\begin{table}[htb]
    \centering
    \caption{Comparison of Inventory Metrics Across Different Methods}
    \label{tab:inventory_method_comparison}
    \begin{tabular}{lccccc}
        \toprule  % 顶部粗线
        Method       & Turnover Days   & In-stock Rate & Holding Cost & Stockout Cost & Total Cost   \\
        \midrule  % 中间细线
        JD online & $X$        & $Y$           & $C_0$        & $C_1$         & $C$          \\
        DL           & $X$-0.8    & $Y$+1.0\%     & -2.9\%      & -14.7\%      & -10.0\%        \\
        OR guided RL          & \bfseries $X$-1.0    & $Y$+0.96\%     & \bfseries -3.67\%      & -14.00\%      & -9.9\%        \\
        OR+expert guided RL & $X$-0.6 & \bfseries $Y$+1.2\% & -1.47\% & \bfseries -17.60\% & \bfseries -11.20\% \\
        \bottomrule  % 底部粗线
    \end{tabular}
\end{table}

Table~\ref{tab:inventory_method_comparison} summarizes the comparison results. Consistent with our analysis, the OR-guided RL model slightly underperforms the DL model in this promotion scenario, confirming that general optimization logic can be overly conservative during extreme events. In contrast, the model augmented with secondary expert fine-tuning effectively overcomes this limitation. By assimilating expert experience, it prioritizes inventory availability and mitigates stockout losses, ultimately yielding an additional 1.2\% reduction in total cost relative to the DL model. These findings empirically validate the necessity of complementing general OR logic with scenario-specific expert insights for critical operations.

\begin{figure}[h]
    \centering
    \caption{replenishment and sales trajectory for a representative SKU during promotion}
    \includegraphics[width=0.8\linewidth]{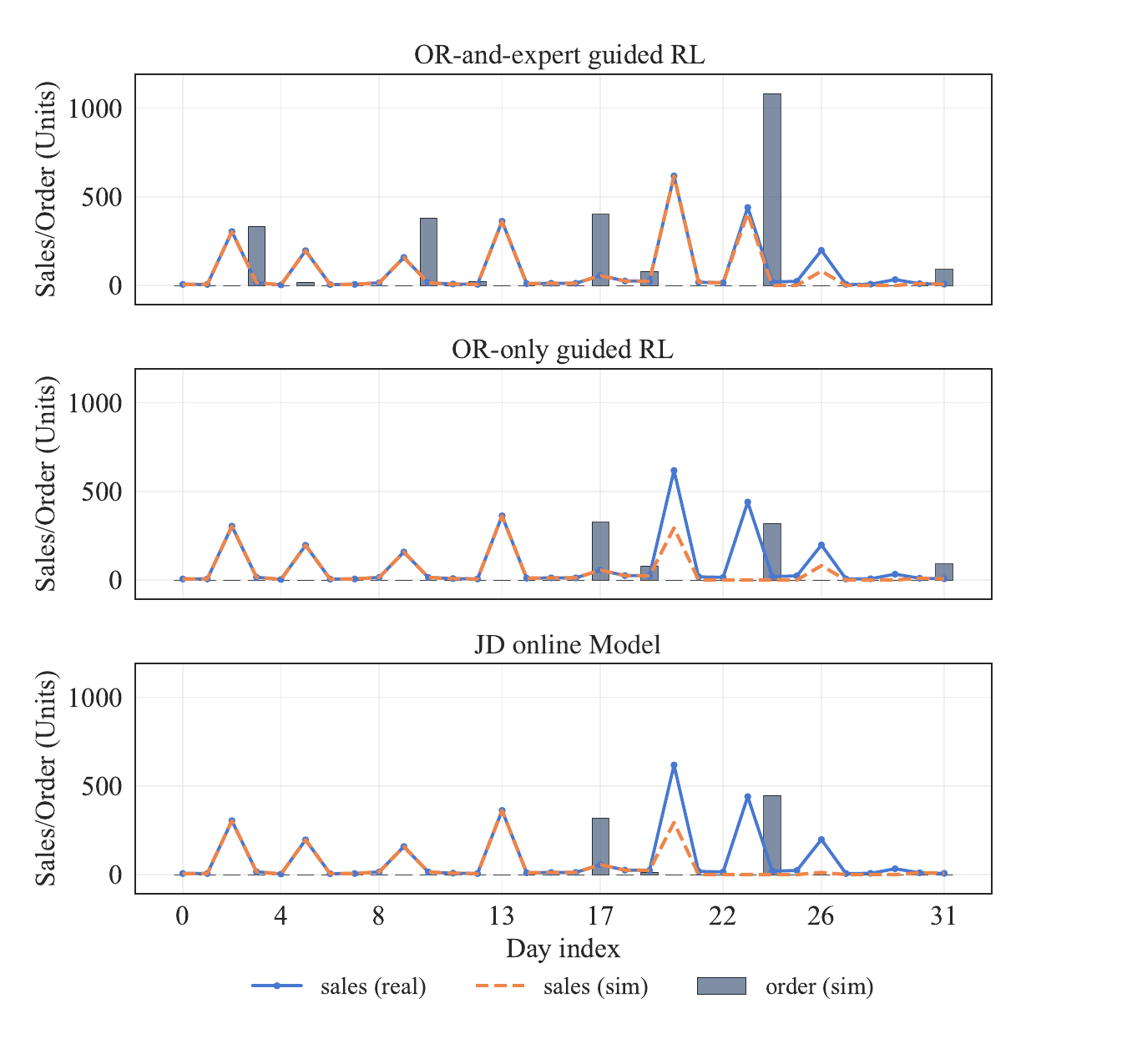}
    \label{fig:rl_sku_case}
\end{figure}

Furthermore, we perform a granular SKU-level analysis to elucidate how different models shape replenishment and sales trajectories. Figure~\ref{fig:rl_sku_case} visualizes these dynamics for a representative item under three distinct models during the 30-day promotional window. The tight overlap between the simulated sales curve and the demand curve indicates that inventory levels are sufficient to satisfy all arriving demand. A critical difference emerges during the demand surge: the expert-augmented model executes proactive inventory buildup, thereby sustaining supply continuity and satisfying peak demand effectively. In contrast, the baseline models fail to pre-stock, leading to a marked divergence between potential and realized sales (i.e., stockouts) during the critical period. Ultimately, the OR-and-expert guided model achieved a 14\% reduction in lost sales, demonstrating effective alignment with the promotional mandate of prioritizing high in-stock rates.

\section{Field Experiment}
\label{sec:field_exp}
Building on the validated success of the offline phase, the project advanced to its critical milestone: live deployment. In September 2025, our Pretrain-then-Reinforce framework has been successfully implemented in the operational system of JD.com to verify its real-world effectiveness. As of October 2025, the model governs the inventory management of \textcolor{blue}{331} SKUs in production, yielding significant cost savings and inventory turnover reduction. In the future, JD.com plans to expand the deployment scope of this framework and explore the potential scaling law by replacing existing lightweight model components with advanced large models, such as the large time-series models.% 加一些后续推广和上线的计划？

In this section, we detail the design and report the results of the \textcolor{blue}{30}-day field experiment conducted within JD.com’s inventory management system from \textcolor{blue}{2025-09-01} to \textcolor{blue}{2025-09-30}.

\subsection{Experiment Design}
To further validate whether our model delivers significant performance improvements in practical operational settings, we conduct a 30-day field experiment in JD.com. These SKUs originated from 3 second-level categories, meat snacks, cakes and pastries, and egg snacks, which belong to the first-level category ``snack foods". 
Regarding parameters, we directly retrieved the holding cost and stockout loss parameter for each SKU from JD.com's internal datasets. 

The treatment group involves 331 SKUs and 3,899 (SKU-DC) pairs. Starting from September 1, 2025, the replenishment decisions for these (SKU, DC) pairs were generated by our proposed model, while the remaining pairs continued to use JD.com's current replenishment algorithm. We collected daily inventory levels, sales volumes, and replenishment orders for all (SKU, DC) pairs during the field experiment to calculate and compare key business metrics for both groups, including holding cost,  inventory turnover, and in-stock rate.

\subsubsection{Experimental Group Design.}
\begin{figure}[h]
    \centering
    \caption{Comparison of the propensity score distribution before and after matching}
    \includegraphics[width=0.9\linewidth]{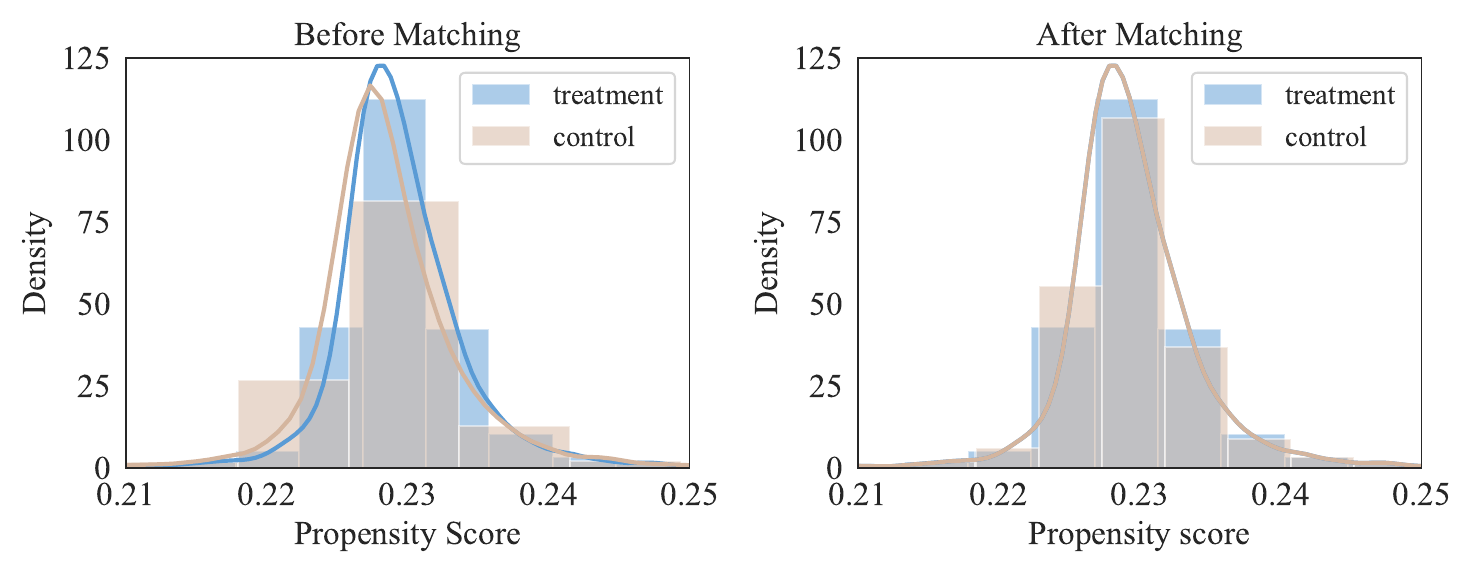}
    \label{fig:psm}
\end{figure}

To thoroughly evaluate the model’s performance under practical operational conditions, the selection of SKUs for the treatment group was determined by JD.com's managers based on specific business criteria, prioritizing SKUs with high automation rates to minimize the confounding effect of human intervention on comparison (the automation rate is defined as the percentage of replenishment decisions that are output by models and adopted for execution, where a higher rate indicates less human interference). To balance the control and treatment groups and achieve near-random assignment, we employed Propensity Score Matching (PSM), selecting the most closely matched SKUs within the corresponding category of each treatment SKU to form the control group, using inventory turnover and demand as confounder variables. 
Figure~\ref{fig:psm} visualizes the propensity score of the control and treatment sets before and after matching, demonstrating the efficacy of this selection methodology.

% 基于PSM挑选对照组&实验组，最近邻挑选最近的一个样本

\subsubsection{Key Performance Indicators (KPIs).}
In addition to the standard cost metrics, we also monitored the \textit{Inventory Turnover Days} and \textit{In-stock Rate} during the evaluation period, which provide a more direct view of operational performance and reveal their trade-off. Since the observed sales directly represent the realized outcomes of the model’s replenishment decisions and the counterfactual outcomes cannot be inferred, the stockout rate and associated losses cannot be accurately evaluated. Therefore, these metrics are excluded from this experiment.
% Specifically, the Turnover Time is evaluated by the average duration a unit of the SKU remained in the warehouse. And the In-Stock Rate is computed as the percentage of days that stockout doesn't occur during the experimental period, measuring the frequency of stockouts.

\subsection{Results}

Table~\ref{tab:inventory_method_comparison_psm} shows the online A/B test performance based on the above propensity score matching. The results indicate that the treatment group exhibits a significant and comprehensive improvement over the control group. 

% 线上效果说明：针对上线品类进行复盘，同品类的商品互相作为对照组，其中高自动化率的SKU作为实验组（认为高自动化的模型补货决策对最后的整体周转&现货指标变化作用较大），其它SKU自动作为对照组。其中，与实际业务复盘指标的经验对齐，考虑到不同sku的周转水平本身有差异（比如高销sku的周转天然就低）、为了对比公平、周转主要看同比变化；而现货率，针对不同sku、我们的理想现货率都是100%、越高越好，所以现货率直接对比绝对值。此外，优于真实情况下缺货损失无法估计，我们仅对比周转变化带来的holding cost变化，可以看到，实验组相对对照组能带来明显的持货成本优化、以及明显的现货率优化
\begin{table}[h]
    \centering
    \caption{Online A/B-test Performance}
    \label{tab:inventory_method_comparison_psm}
    \begin{tabular}{lccccc}
        \toprule  % 顶部粗线
        Groups & Turnover Days & In-stock Rate & Holding Cost \\
        \midrule  % 中间细线
        control group & X & Y & Z \\
        treatment group & X\textbf{-5.27} & Y\textbf{+2.29\%} & \textbf{-29.95\%} \\
        \bottomrule  % 底部粗线
    \end{tabular}
\end{table}

Specifically, our model achieved a reduction of \textbf{5.27} turnover days and a \textbf{29.95\%} decrease in holding costs, while simultaneously increasing the in-stock rate by \textbf{2.29\%}. In the context of large-scale retail operations, typically characterized by a rigid trade-off between service levels and inventory costs, these results are particularly notable. 
They suggest that our framework enables the retailer to maintain higher service levels with significantly less capital occupation, validating its ability to execute precise replenishment decisions aligned with actual demand rhythms.

\subsection{DiD}
To further mitigate the confounding effect of time on product sales and to ensure comparability, we also incorporated a $T=30$ day period from the corresponding time window of the previous year (September 1, 2024 to September 30, 2024). We then compared the performance of this pre-experiment period against the performance of both the treatment group and the control group during the actual experiment period (September 1, 2025 to September 30, 2025). During the pre-experiment period, both the treatment and control groups operated using JD.com’s current algorithm. Table~\ref{tab:did_results} presents the results of the Difference-in-Differences (DiD) comparison.

\begin{table}[h]
    \centering
    \caption{Online Performance (DID)}
    \label{tab:did_results}
    \begin{tabular}{lccccc}
        \toprule  % 顶部粗线
        Groups & Turnover Days & In-stock Rate & Holding Cost (year on year) \\
        \midrule  % 中间细线
        control group & X & Y & +22.05\% \\
        treatment group & \textbf{X-4.07} & \textbf{Y+0.47\%} & \textbf{+3.7\%} \\
        \bottomrule  % 底部粗线
    \end{tabular}
\end{table}

The DiD analysis effectively isolates the model's contribution by treating the control group's year-on-year variation as the baseline for external market trends. As shown in the table, the control group experienced a significant \textbf{22.05\%} surge in holding costs, likely driven by factors such as market prosperity or seasonal demand expansion. Against this backdrop of rising costs, our model demonstrated exceptional control. It limited the holding cost increase to a mere \textbf{3.7\%}—effectively achieving a relative reduction of \textbf{18.35\%} compared to the baseline trend exhibited by JD.com's incumbent algorithm. Crucially, this cost efficiency was not achieved at the expense of operational quality; on the contrary, the treatment group simultaneously outperformed the control trend by reducing turnover days by \textbf{4.07} and increasing the in-stock rate by \textbf{0.47\%}.

This analysis confirms that the performance gains observed in the field experiment are causal and intrinsic to our proposed framework, rather than artifacts of favorable seasonal trends. By effectively decoupling inventory performance from adverse external volatility, our model demonstrates superior resilience and reliability for real-world deployment.
\section{Conclusion}
\label{sec:conclusion}

In this work, we propose a novel \textit{OR-Guided ``Pretrain-then-Reinforce" }framework for inventory management. Revisiting the two foundational questions posed in the introduction regarding the synergy of AI and OR, our study offers a concrete realization of the ideal integration. We demonstrate that: (i) the distinct roles are best characterized with AI as the engine for adaptive perception and OR as the provider of structural logic; and (ii) this synergy is most effectively harmonized not through sequential pipelines, but through a deep alignment architecture. By leveraging the structured guidance of a simulation-augmented OR model alongside the critical alignment-exploration mechanism of RL, our framework enables the AI agent to internalize the complex economic trade-off between inventory turnover and stockout losses, effectively grounding its decision logic in rigorous operational principles. 

Crucially, this successful practice challenges the prevailing reliance on brute-force model scaling. We demonstrate that structured OR guidance combined with our domain-informed ``Pretrain-then-Reinforce" architecture can serve as an effective substitute for massive parameter scales, delivering state-of-the-art performance and strong adaptability in dynamic operational contexts. Validated through a comprehensive suite of numerical experiments and a successful field deployment at JD.com, our framework substantially outperforms both traditional heuristics and incumbent enterprise practices, providing a scalable and cost-effective solution for real-world supply chains.

Future research may expand on this work in several promising directions. From a modeling perspective, integrating our framework with larger foundation models, such as the JD.com’s large time-series model, enables the exploration of scaling laws in decision-making tasks. Another vital direction lies in generalizing this paradigm to more complex inventory structures, such as multi-echelon systems and inventory allocation problems, where the synergy of AI and OR could unlock even greater efficiency.
%\input{files/appendices}

%\THEEndNotes
%\begingroup \parindent 0pt \parskip 0.0ex \def\enotesize{\normalsize} \theendnotes \endgroup

% Acknowledgments here
% \ACKNOWLEDGMENT{We would like to express our sincere gratitude to [acknowledge individuals, organizations, or institutions] for their invaluable contributions to this research. We are also grateful to [mention any additional acknowledgements, such as technical assistance, data providers, or colleagues] for their support and assistance throughout the course of this work.}

% References here (outcomment the appropriate case)

% CASE 1: BiBTeX used to constantly update the references
%   (while the paper is being written).
%\bibliographystyle{informs2014} % outcomment this and next line in Case 1
%\bibliography{<your bib file(s)>} % if more than one, comma separated

%\bibliographystyle{informs2014} % outcomment this and next line in Case 1
%\bibliography{sample} % if more than one, comma separated

% CASE 2: BiBTeX used to generate mypaper.bbl (to be further fine tuned)
%\input{mypaper.bbl} % outcomment this line in Case 2

%If you don't use BiBTex, you can manually itemize references as shown below.

%\bibliographystyle{nonumber}
\bibliographystyle{informs2014} 
\bibliography{main} 

% Appendix here
\begin{APPENDICES}
\newpage
\section{Implementation Details \& Hyperparameters}
All models in our framework are implemented using the PyTorch framework. A distinct advantage of our domain-informed architecture is its computational efficiency. Unlike large-scale foundation models that necessitate massive cluster resources, our entire pipeline, which encompasses both pre-training and reinforcement learning fine-tuning, was successfully executed on a single general-purpose data center GPU, 10 CPU cores and 40G RAM. The pre-training phase typically converges in approximately 1.5-2 hours, while the RLOO fine-tuning stage requires only 5-6 hours, highlighting the framework's accessibility and low resource footprint.
\subsection{Deep Neural Network Architecture Details}
Our model architecture is designed to balance expressiveness with parameter efficiency, avoiding unnecessary complexity while ensuring sufficient capacity to capture temporal dynamics. Table~\ref{tab:dnn_hyperparams} provides a comprehensive summary of the structural hyperparameters.

\paragraph{Feature Extraction Module.} 
The backbone of our model is a multi-path Transformer encoder designed to distill temporal dependencies from heterogeneous data streams. We project the raw inputs into a unified embedding dimension of $d_{model}=128$. To effectively capture sequential patterns, we utilize a stack of 2 Transformer encoder layers, each equipped with 4 multi-head self-attention mechanisms. To prevent overfitting given the data scale, we apply a dropout rate of 0.1 across the attention mechanisms and positional feed-forward networks. Finally, the encoded outputs from these three distinct feature streams are directly concatenated to construct the final fused temporal feature embedding.

\paragraph{Multi-Objective Prediction Heads.} 
The learned feature representation serves as the input for two distinct heads. The \textit{Sales Forecasting Head} is implemented as a straightforward Multi-Layer Perceptron (MLP) with hidden dimensions of [128, 64, 32], activated by ReLU functions, focusing on accurate demand sensing. The \textit{Replenishment Decision Head}, the core decision-making unit, employs a Variational Autoencoder (VAE) structure to model the stochasticity of optimal decisions. The encoder projects the feature embedding into a latent space of dimension $z_{dim}=32$, from which the decoder generates the unnormalized logits for the replenishment action $a$, preserving the full distributional information for downstream RL.

\paragraph{Optimization and Training Setup.}
Regarding the optimization process, we train the model using the Adam optimizer. The learning rate is initialized at 1e-4 and adjusted via a CosineAnnealingLR scheduler to ensure stable convergence. We utilize a batch size of 32-64 to balance memory efficiency and gradient estimation accuracy during the pre-training phase.

\begin{table}[h]
    \centering
    \small
    \caption{Hyperparameters of the Deep Neural Network Architecture}
    \label{tab:dnn_hyperparams}
    \renewcommand{\arraystretch}{1.2}
    \begin{tabular}{l|c}
    \toprule
    \textbf{Hyperparameter} & \textbf{Value} \\
    \midrule
    \multicolumn{2}{l}{\textit{\textbf{Feature Extraction (Transformer)}}} \\
    \midrule
    Embedding Dimension ($d_{model}$) & 128 \\
    Number of Encoder Layers & 2 \\
    Number of Attention Heads & 4 \\
    Feed-forward Hidden Dimension & 256 \\
    Dropout Ratio & 0.1 \\
    \midrule
    \multicolumn{2}{l}{\textit{\textbf{Sales Forecasting Head (MLP)}}} \\
    \midrule
    Number of Layers & 3 \\
    Hidden Dimensions & [128, 64, 32] \\
    Activation Function & ReLU \\
    \midrule
    \multicolumn{2}{l}{\textit{\textbf{Replenishment Decision Head (VAE)}}} \\
    \midrule
    Encoder Architecture & [512, 256, 2*$z_{dim}$] \\ 
    Decoder Architecture & [$z_{dim}$,256,512]\\
    Latent Dimension ($z_{dim}$) & 32 \\
    $\beta$ (KL Weight in ELBO) & 1 \\
    \midrule
    \multicolumn{2}{l}{\textit{\textbf{Optimization (Pre-training)}}} \\
    \midrule
    Optimizer & Adam \\
    Learning Rate & 1e-4 \\
    Batch Size & 32 \\
    Learning Rate Scheduler & CosineAnnealingLR \\
    \bottomrule
    \end{tabular}
\end{table}

\subsection{RLOO Training Configuration}
The reinforcement learning stage focuses on aligning the pre-trained policy with OR-guided logic and expert preferences. This process is governed by the specific configuration of the Reinforcement Leave-One-Out (RLOO) algorithm \citep{ahmadian2024back}, as detailed in Table~\ref{tab:rloo_hyperparams}.

\paragraph{Alignment and Exploration.} 
To ensure the policy effectively internalizes OR logic without deviating excessively from the pre-trained baseline, we set the KL-divergence penalty coefficient $\beta$ to $0.1$. The hybrid reward function balances the rigorous rule-based alignment and flexible simulation-based exploration via the weight parameter $\omega = 0.5$. During each training step, we generate $k=10$ independent samples per prompt to estimate the leave-one-out baseline. This strategy provides a low-variance gradient estimate, enabling stable and efficient policy updates even with a limited simulation budget.

\begin{table}[h]
    \centering
    \small
    \caption{Hyperparameters for RLOO Fine-Tuning}
    \label{tab:rloo_hyperparams}
    \renewcommand{\arraystretch}{1.2}
    \begin{tabular}{lcl}
    \toprule
    \textbf{Parameter} & \textbf{Value} & \textbf{Description} \\
    \midrule
    $k$ (Samples) & 10 & Number of generations per input for baseline estimation \\
    $\beta$ (KL Coeff.) & 0.1 & Penalty weight for deviation from reference policy \\
    $\omega$ (Reward Weight) & 0.5 & Balance between rule-based and simulation rewards \\
    Learning Rate & 1e-4 & Step size for parameter updates \\
    Batch Size & 32 & Number of decision instances per update step \\
    Gradient Accumulation & 1 & Steps to accumulate gradients before update \\
    \bottomrule
    \end{tabular}
\end{table}

\subsection{Simulator Details: High-Fidelity Trace-Driven Evaluation Engine}
\label{app:simulator}
\paragraph{Industrial-Grade Infrastructure.} 
Our simulation environment is built upon the foundational infrastructure of JD.com’s intelligent supply chain platform. JD.com has established an \textit{OR+AI accelerated end-to-end simulation system} with ultra-high resolution, capable of digitally replicating distribution networks, procurement, replenishment, and fulfillment at the granularity of individual orders. While the full-scale system handles complex multi-echelon network design involving tens of millions of SKUs, our study extracts a specialized, lightweight \textit{replenishment evaluation module} from this ecosystem. This ensures that although our simulation logic is streamlined to focus on the replenishment decision node, the high fidelity and industrial precision of the parent system are maintained through the strict incorporation of granular input parameters, such as holding costs, dynamic lead times, and SKU-specific profit margins.

\paragraph{Trace-Driven Mechanism.} 
To avoid the model misspecification risks inherent in parametric stochastic modeling, we adopt a \textit{trace-driven simulation approach}. The simulator "replays" historical demand and arrival sequences exactly as they occurred. This allows us to implicitly capture complex, non-stationary market dynamics, including supply shocks and the pulse of promotional events (e.g., Double 11 in China), which are often lost in simplified probability distributions. 
To ensure strict adherence to real-world operational environments, the simulation explicitly incorporates rigorous business constraints and granular parameters, including \textit{Nominal Review Time (NRT), dynamic Vendor Lead Time (VLT), Minimum Order Quantities (MOQ), and SKU-specific cost structures}. The state transition logic follows the standard discrete-event sequence:
\begin{enumerate}
    \item \textbf{Demand Fulfillment:} Realized demand is deducted, and lost sales are recorded only when inventory is insufficient, reflecting the ``lost-sales" reality of e-commerce. Crucially, this fulfillment process preserves the \emph{topological fidelity} of the actual fulfillment network (e.g., the 2-tier RF structure in Figure~\ref{fig:RF_instance}). It enforces strict regional availability constraints: sales are lost if the local node chain (e.g., F1 and R) is depleted, regardless of inventory in parallel sub-networks.
    \item \textbf{Arrival Processing:} Scheduled shipments arrive and update on-hand inventory, strictly adhering to VLT delays.
    \item \textbf{Metric Accumulation:} Daily holding costs and stockout penalties are aggregated based on real-time SKU attributes (price, margin, warehouse fees).
    \item \textbf{Replenishment Trigger:} Counterfactual orders are generated based on the candidate decision $v$ in accordance with the actual review schedule of the SKU.
\end{enumerate}
\vspace{-0.4em}
\begin{figure}[htb]
    \centering
    \caption{Logic flow of a representative 2-tier RF fulfillment instance}
    \includegraphics[width=0.6\linewidth]{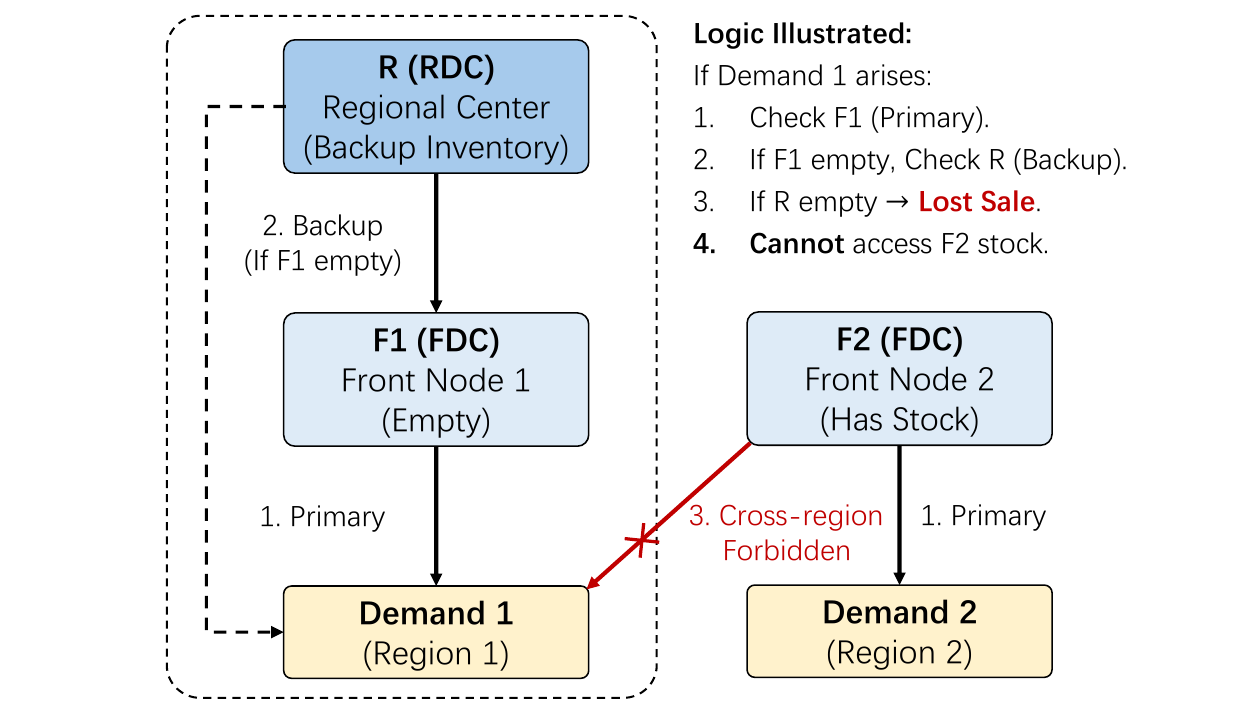}
    \label{fig:RF_instance}
\end{figure}
\vspace{-1em}
\paragraph{Matrix-Based Parallelization for Exhaustive Evaluation.} 
A key computational challenge lies in the label generation phase prior to training. To derive the OR-optimal labels, we must perform an exhaustive counterfactual evaluation for every decision instance in the training set (i.e., every SKU at every decision point). This entails assessing the outcomes of \textit{all} possible integer decision values $v \in [L, U]$ over a horizon $T$ to provide the necessary inputs for our optimization model and the subsequent turnover-based calibration. A traditional sequential simulation would be computationally prohibitive given the scale. 
To address this, we implemented a fully vectorized simulation engine. By transforming the logic into matrix operations, we evaluate thousands of (SKU-decision) combinations in parallel. This design choice prioritizes computational throughput without sacrificing empirical fidelity, enabling the rapid generation of the large-scale, physics-compliant labels required for our framework.

\paragraph{Interactive Framework Between Replenishment Simulator and RLOO Model.}
To establish a tightly coupled integration between the replenishment simulation module and our RL optimization algorithm (i.e., RLOO), we design a standardized interactive framework. This framework connects decision generation, scenario simulation, and feedback optimization through three core implementation mechanisms:
\begin{itemize}
    \item \textbf{MDP-Compatible Interface:} The simulator is encapsulated as an RL-compatible environment class adhering to the Markov Decision Process (MDP). It receives the core replenishment action from the RLOO policy and converts the target inventory adjustment into a concrete target inventory level contextualized by forecasted demand. After executing the full simulation logic, the environment outputs key business metrics including inventory turnover rate, inventory holding costs, and stockout penalty costs, which serve as the critical feedback signal for the RLOO algorithm to update policy parameters.
    \item \textbf{State Fidelity and Pipeline Dynamics:} To ensure decision validity, each simulation episode initializes the inventory state using actual historical on-hand data. Furthermore, the simulator rigorously accounts for pipeline dynamics, tracking both existing in-transit orders (based on scheduled arrivals) and new orders triggered by the RL agent. This ensures that simulated inventory fluctuations align precisely with the rhythmic reality of supply chain operations.
    \item \textbf{Strategic Simulation Horizon:} We specifically calibrate the simulation horizon to capture the long-term consequences of immediate decisions. Given an average supplier lead time of approximately 7 days, we set the simulation length to 31 days. This duration extends 10 to 20 days beyond the expected arrival of the current replenishment order, ensuring that the simulator fully captures the cascading impacts of immediate decisions on future costs and fulfillment. Consequently, this provides the RLOO algorithm with comprehensive, long-term feedback for policy optimization.
\end{itemize}

\end{APPENDICES}
% Options are (1) APPENDIX (with or without general title) or
%             (2) APPENDICES (if it has more than one unrelated sections)
% Outcomment the appropriate case if necessary
%
% \begin{APPENDIX}{<Title of the Appendix>}
% \end{APPENDIX}
%
%   or
%
% \begin{APPENDICES}
% \section{<Title of Section A>}
% \section{<Title of Section B>}
% etc
% \end{APPENDICES}

%%%%%%%%%%%%%%%%%
\end{document}